\documentclass[final]{cvpr}

\usepackage{times}
\usepackage{epsfig}
\usepackage{graphicx}
\usepackage{amsmath}
\usepackage{amssymb}
\usepackage[pagebackref=true,breaklinks=true,colorlinks,bookmarks=false]{hyperref}

\def\cvprPaperID{543} % *** Enter the CVPR Paper ID here
\def\confYear{CVPR 2021}

\def\wrt{w.r.t\onedot} 
\usepackage{multirow}
\usepackage{booktabs}
\usepackage{subfigure}
\usepackage{amsmath,amsfonts,bm}

\newcommand{\Cell}{Searching block}
\newcommand{\cell}{searching block}
\newcommand{\cells}{searching blocks}
\newcommand{\alias}{HR-NAS}
\usepackage{appendix}

\def\eqref#1{equation~\ref{#1}}
\def\1{\bm{1}}

\DeclareMathAlphabet{\mathsfit}{\encodingdefault}{\sfdefault}{m}{sl}
\SetMathAlphabet{\mathsfit}{bold}{\encodingdefault}{\sfdefault}{bx}{n}

\newcommand{\softmax}{\mathrm{softmax}}

\begin{document}

%%%%%%%%% TITLE
\title{HR-NAS: Searching Efficient High-Resolution Neural Architectures \\ with Lightweight Transformers}

\author{
  Mingyu Ding\textsuperscript{\rm 1}\thanks{This work was done as a part of internship at Bytedance AI Lab US.}~~
  Xiaochen Lian\textsuperscript{\rm 2}~~
  Linjie Yang\textsuperscript{\rm 2}~~
  Peng Wang\textsuperscript{\rm 2}~~
  Xiaojie Jin\textsuperscript{\rm 2}~~
  Zhiwu Lu\textsuperscript{\rm 3}~~
  Ping Luo\textsuperscript{\rm 1}~~\\
 \textsuperscript{\rm 1}The University of Hong Kong~~~~~~~~~~~
 \textsuperscript{\rm 2}Bytedance Inc.~~~~~~~~~~~\\
 \textsuperscript{\rm 3}Gaoling School of Artificial Intelligence, Renmin University of China\\
 \texttt{\normalsize\{myding, pluo\}@cs.hku.hk~~~~~~~~
 \texttt{\normalsize luzhiwu@ruc.edu.cn}}\\
 \texttt{\normalsize\{xiaochen.lian, linjie.yang, peng.wang, jinxiaojie\}@bytedance.com}
}

\maketitle
\pagestyle{empty}  % no page number for the second and the later pages
\thispagestyle{empty}

%%%%%%%%% ABSTRACT
\begin{abstract}

High-resolution representations (HR) are essential for dense prediction tasks such as segmentation, detection, and pose estimation. Learning HR representations is typically ignored in previous Neural Architecture Search (NAS) methods that focus on image classification. This work proposes a novel NAS method, called HR-NAS, which is able to find efficient and accurate networks for different tasks, by effectively encoding multiscale contextual information while maintaining high-resolution representations. In HR-NAS, we renovate the NAS search space as well as its searching strategy. To better encode multiscale image contexts in the search space of HR-NAS, we first carefully design a lightweight transformer, whose computational complexity can be dynamically changed with respect to different objective functions and computation budgets. To maintain high-resolution representations of the learned networks, HR-NAS adopts a multi-branch architecture that provides convolutional encoding of multiple feature resolutions, inspired by HRNet~\cite{wang2020deep}. Last, we proposed an efficient fine-grained search strategy to train HR-NAS, which effectively explores the search space, and finds optimal architectures given various tasks and computation resources. As shown in Fig.\ref{fig:comparison} (a), HR-NAS is capable of achieving state-of-the-art trade-offs between performance and FLOPs for three dense prediction tasks and an image classification task, given only small computational budgets. For example, HR-NAS surpasses SqueezeNAS~\cite{shaw2019squeezenas} that is specially designed for semantic segmentation while improving efficiency by 45.9\%. Code is available at \href{https://github.com/dingmyu/HR-NAS}{https://github.com/dingmyu/HR-NAS}.

\end{abstract}

\vspace{-12pt}
\section{Introduction}

\begin{figure}[t]
  \centering
%   (a)
    \begin{minipage}[t]{0.49\linewidth}
      \centering
      \includegraphics[width=\linewidth]{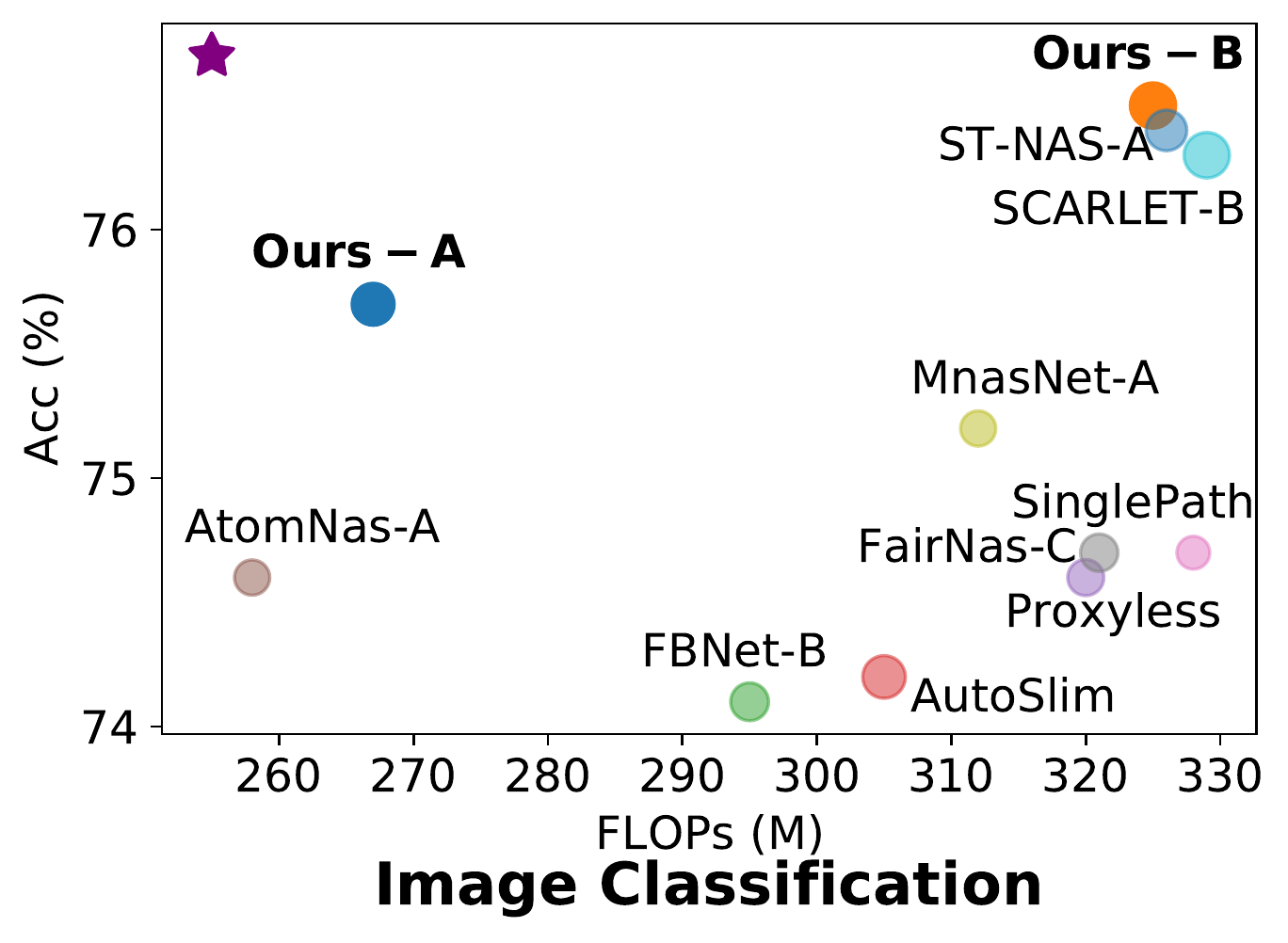}\vspace{0pt}
      \includegraphics[width=\linewidth]{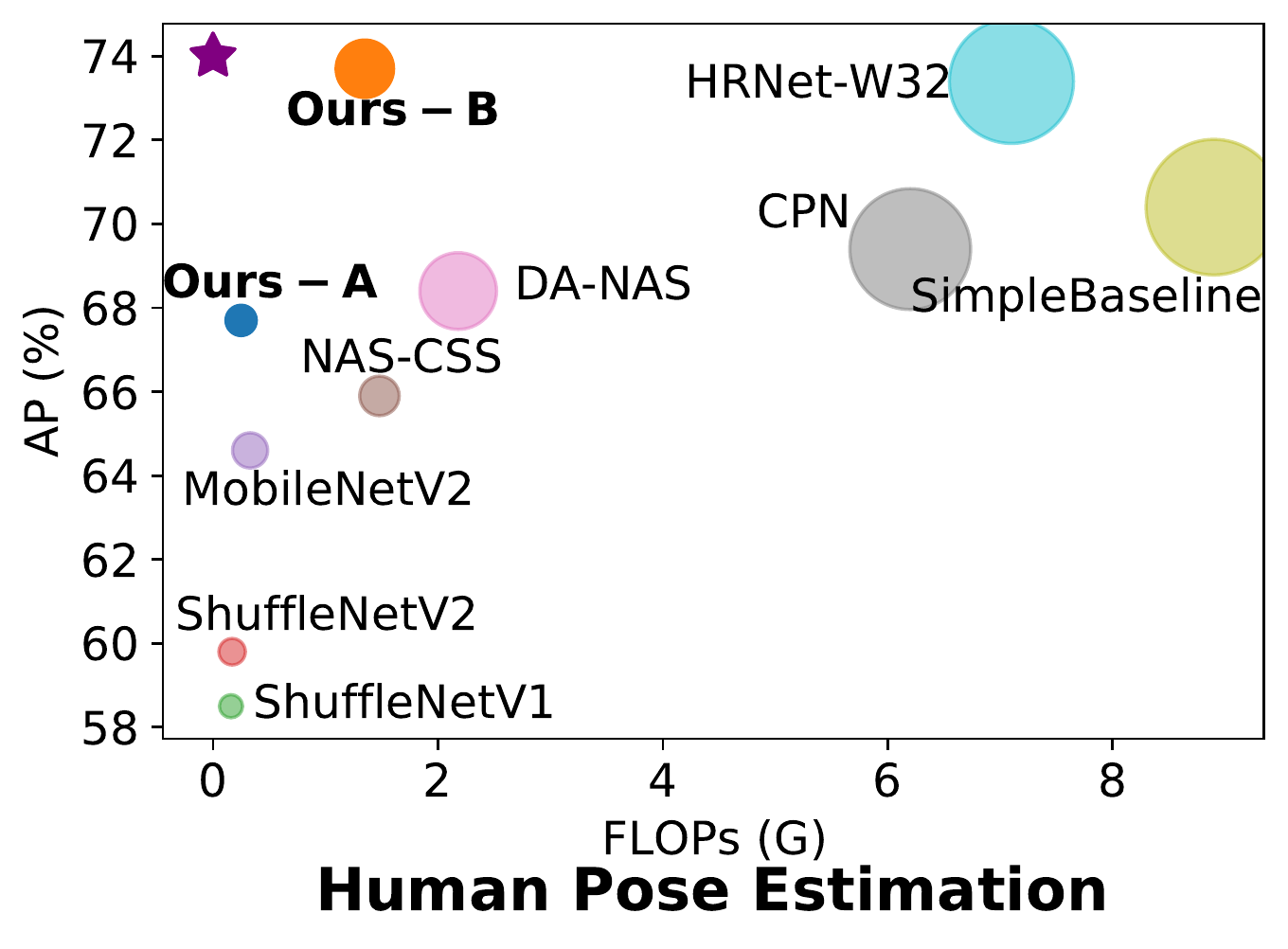}\vspace{0pt}
    \end{minipage}
  \hfill
    \begin{minipage}[t]{0.49\linewidth}
      \centering
      \includegraphics[width=\linewidth]{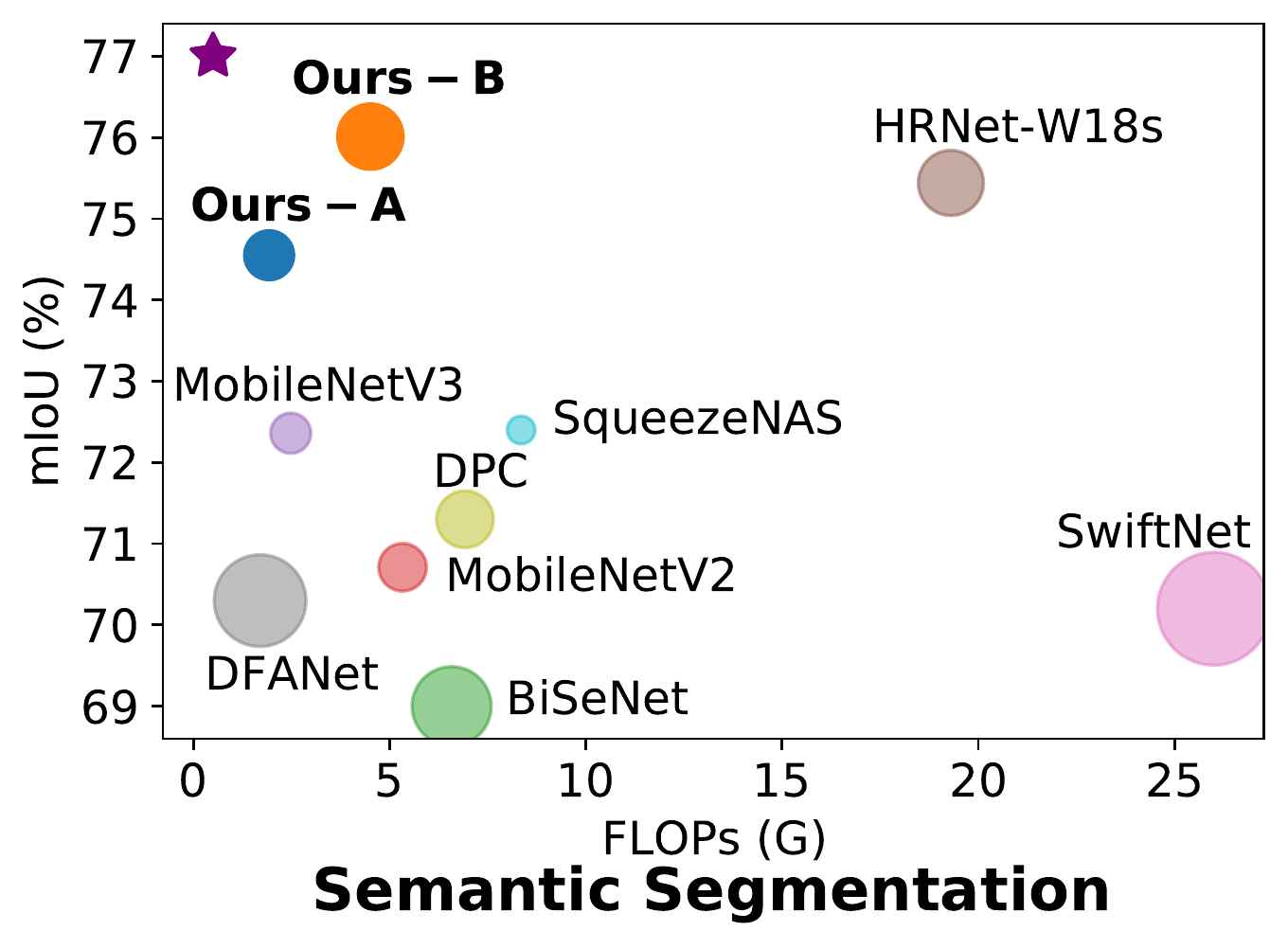}\vspace{0pt}
      \includegraphics[width=\linewidth]{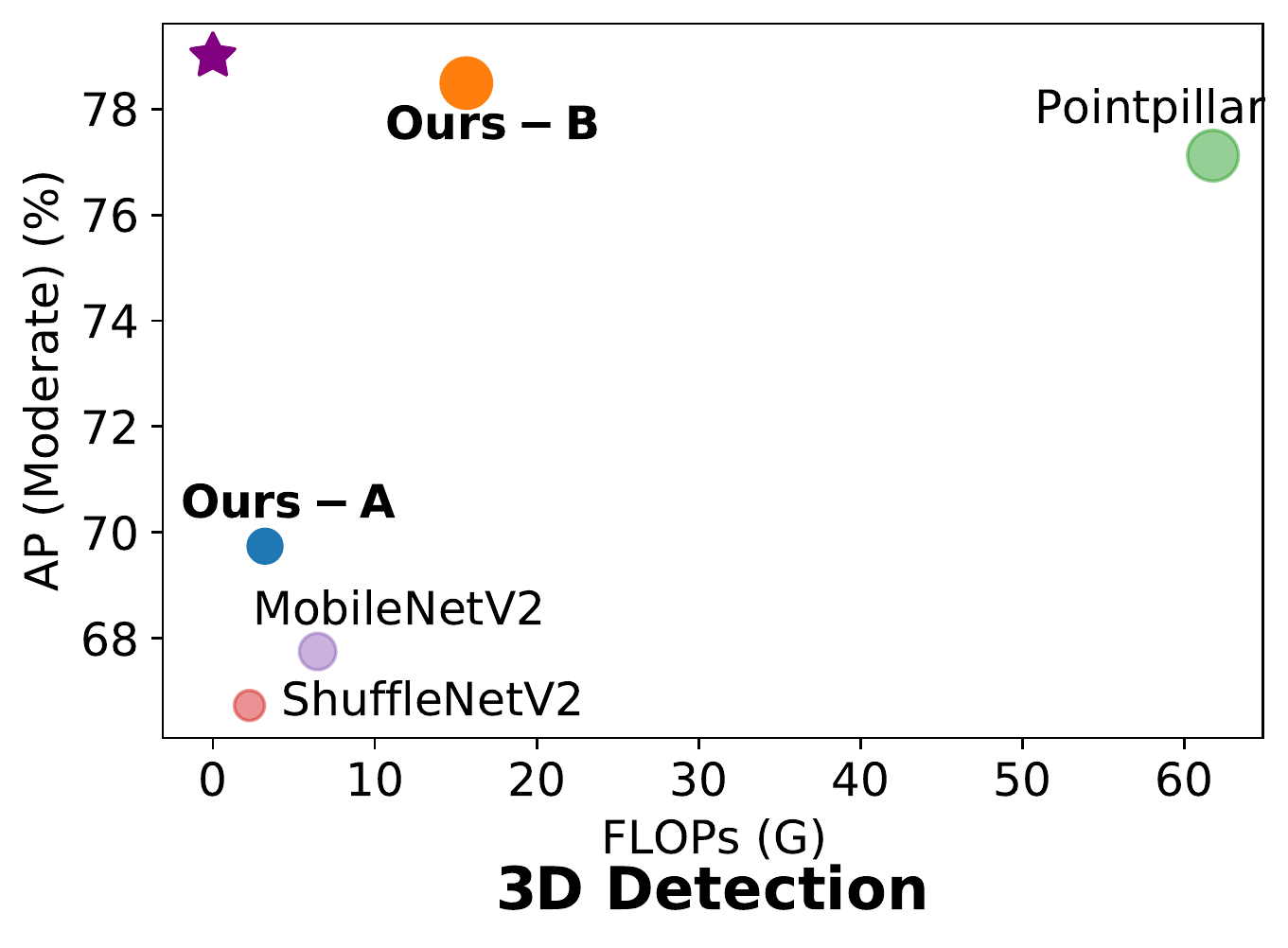}\vspace{0pt}
    \end{minipage}
%   \hfill
    % \begin{minipage}[b]{0.05\linewidth}
    % (b) \\~\\~\\
    % \end{minipage}
    % \begin{minipage}[t]{0.94\linewidth}
    %   \centering
    %   \includegraphics[width=\linewidth]{figures/query-resolution.pdf}\vspace{0pt}
    % \end{minipage}
  \caption{Comparisons of the efficiency (\ie, FLOPs) and the performance (\eg, Acc, mIoU, AP) on 4 computer vision tasks, \ie, classification (ImageNet), segmentation (CityScapes), pose estimation (COCO), and 3D detection (KITTI),
  between the proposed approach and existing SoTA methods. Each method is represented by a circle, whose size represents the number of parameters. $\bigstar$ represents the optimal model with both high performance and low FLOPs. Our approach achieves superior performance under similar FLOPs compared to its counterparts on all four benchmarks.
%   (b) The average number of queries for our transformers of different resolutions of the searched model on four tasks. Our model adaptively allocates the computational budget to tasks with different preferences.
  }
  \label{fig:comparison}
  \vspace{-8pt}
\end{figure}

Neural architecture search (NAS) has achieved remarkable success in automatically designing efficient models for image classification \cite{wu2019fbnet,liang2019darts+,hanxiao2019darts,mei2020atomnas,han2019proxyless,xie2018snas,cai2019once,shaw2019meta, zhou2019epnas, fang2020densely}. 
NAS has also been applied to improve the efficiency of models for dense prediction tasks such as semantic segmentation~\cite{shaw2019squeezenas,chen2018searching} and pose estimation~\cite{dai2020data}. 
However, existing NAS methods for dense prediction either directly extend the search space designed for image classification~\cite{dai2020data,lin2020graph}, only search for a feature aggregation head~\cite{nekrasov2019fast,chen2018searching}, organizing network cells in a chain-like single-branch manner~\cite{liu2019auto,shaw2019squeezenas}. 
This lack of consideration to the specificity of dense prediction hinders the performance advancement of NAS methods compared to the best hand-crafted models~\cite{wang2020deep,ding2021learning}.

In principle, dense prediction tasks require the integrity of the global context and the high-resolution (HR) representation; the former is critical to clarify ambiguous local features~\cite{zhang2018context} at each pixel, and the latter is useful for the accurate prediction of fine details~\cite{kirillov2020pointrend}, such as semantic boundaries and keypoint locations. However, these two aspects, especially the HR representations, have not got enough attention in existing NAS algorithms for classification. 
The straightforward strategy to implement the principle is manually combining multi-scale features at the end of the network~\cite{liu2019auto,chen2017deeplab,li2019dfanet}, while recent approaches~\cite{ding2021learning,wang2020deep,zhang2020dcnas} show the performance can be enhanced by putting multi-scale feature processing within the network backbone. 
Another observation from recent research is that multi-scale convolutional representations can not guarantee a global outlook of the image since dense prediction tasks often come with high input resolution but a network often only covers a fixed receptive field.
Therefore, global attention strategies such as SENet~\cite{hu2018squeeze} and non-local network~\cite{wang2018non} have been proposed to enrich image convolutional features. Most recently, inspired by its success in natural language processing, Transformer architectures~\cite{vaswani2017attention,so2019evolved}, which contain global attention with spatial encoding, have also shown superior results when combined with convolutional neural network for image classification~\cite{anonymous2021an} and object detection~\cite{carion2020end}.

Motivated by the above observations, in this work, we propose a NAS algorithm, which incorporates these strategies, \ie in-network multi-scale features and transformers, and enables their adaptive changing with respect to task objectives and resource constraints. In practice, it is non-trivial to put them together. Firstly, Transformer has a high computational cost that is quadratic \wrt image pixels and hence unfriendly to the NAS search space of efficient architectures. We solve this through a dynamic down projection strategy, yielding a lightweight and plug-and-play transformer architecture that can be combined with other convolutional neural architectures. 
In addition, searching a fused space of multi-scale convolution and transformers needs proper feature normalization, selection of fusion strategies and balancing. We did extensive studies to calibrate various model choices that generalize to multiple tasks. 
% Fig.~\ref{fig:comparison} (b) shows the preferences of different tasks for the number of queries of the transformer.

% In summary, our NAS algorithm, namely HR-NAS, works as follows.  
In summary, HR-NAS works as follows. We first setup a super network, where each layer contains a multi-branch parallel module followed by a fusion module. The parallel module contains searching blocks with multiple resolutions, and the fusion module contains searching blocks of feature fusion determining how feature from different resolutions fuses. Then, based on the computational budget and the task objective, a fine-grained progressive shrinking search strategy is introduced to prune redundant channels in convolutions and queries in transformers, resulting in an efficient model that provides the best trade-off between performance and computational costs. 
With extensive experiments, HR-NAS achieves state-of-the-art on multiple dense prediction tasks and competitive results on image classification under highly efficient settings with a single search. 
Fig.~\ref{fig:comparison} shows a comprehensive comparison of our proposed approach with previous NAS approaches as well as manually designed networks on four different tasks. %We will release our codes and models upon the publication of this paper.

Our main \textbf{contributions} are three-fold.
(1) We introduce a novel lightweight and plug-and-play transformer, which is highly efficient and can be easily combined with convolutional networks for computer vision tasks.
(2) We propose a well-designed multi-resolution search space containing both convolutions and transformers to model in-network multi-scale information and global contexts for dense prediction tasks. To our best knowledge, we are the first to integrate transformers in a resource-constrained NAS search space for computer vision.
(3) A resource-aware search strategy allows us to customize efficient architectures for different tasks.
Extensive experiments show models produced by our NAS algorithm achieve state-of-the-art on three dense prediction tasks and four widely used benchmarks with lower computational costs.
\section{Related Work}

\noindent\textbf{Transformers.}
Transformer~\cite{vaswani2017attention,so2019evolved}, a model architecture relying on a self-attention mechanism to learn dependencies between input and target, is used primarily in natural language processing.
Generative Pre-trained Transformer (GPT) uses language modeling as a pre-training task~\cite{radford2018improving,brown2020language}.
BERT~\cite{devlin2018bert} improves Transformer with a masked language model and a learned positional embedding to replace the sinusoidal positional encoding~\cite{vaswani2017attention}.
%

%modeling 
Since Transformer is suitable for capturing global information and pairwise interactions, some attempts~\cite{wang2018non,carion2020end,anonymous2021an,wu2020visual} have been made to adapt it to computer vision.
%Most existing methods~\cite{hu2018squeeze,hu2018gather,woo2018cbam,wu2019squeezesegv2,bello2019attention,zhao2020exploring,hu2019local} simply apply self-attention to images by using either channel-wise or spatial-wise attention to obtain the per-channel or per-pixel importance and multiplying the attention to the feature map.
%The simplified self-attention lose the spatial and sequential modeling capabilities of the Transformer.
Non-local networks~\cite{wang2018non} proposed a self-attention architecture to capture long-range interactions which can be viewed as a simplified version of Transformer.
DETR~\cite{carion2020end} formulates object detection as a set prediction problem, which is naturally modeled as a sequence prediction task by the Transformer.
%Image Transformer~\cite{parmar2018image} generalizes the Transformer to a local sequence modeling formulation of image generation with a tractable likelihood.
Visual Transformers~\cite{wu2020visual} represent images as a set of visual tokens and apply a Transformer-based structure to detect relationships between visual semantic concepts for semantic segmentation.
iGPT~\cite{chen2020generative} uses a standard Transformer to unsupervisedly learn generative relationships of image pixels.
However, since its computational complexity grows quadratically with the number of pixels, such applications of Transformers in computer vision are computationally expensive.
Some approaches~\cite{ren2020distill,liu2020fastbert,jiao2019tinybert,wang2020linformer,child2019generating} leverage network compression techniques, such as dynamic routing and knowledge distillation, to improve the efficiency of Transformers in NLP. However, efficient Transformers are seldom explored in computer vision.
%
%Moreover, all existing methods adopting Transformers for computer vision\cite{carion2020end,wu2020visual,anonymous2021an,chen2020generative} specify a fixed number of queries and are always customized for specific tasks, making them redundant and non-generalizable.
In light of this, we formulate Transformer into an efficient and plug-and-play module that is seamlessly integrated into a well-designed NAS search space.
% \LXC{we design an efficient and plug-and-play transformer which is seamlessly integrated into a well-designed NAS search space}. The number of queries in our lightweight Transformer is configurable, which can be jointly optimized during an end-to-end structure search, facilitating a flexible allocation of computational cost for various tasks.

\noindent\textbf{Neural Architecture Search for Efficient models.}
Early approaches utilize reinforcement learning~\cite{zoph2017nasnet} and evolution algorithms~\cite{real2017large,lu2018nsga} to find efficient and powerful network structures. However, these methods are usually computationally expensive.
To improve the efficiency of the search process, differentiable search methods such as Darts~\cite{hanxiao2019darts,jin2019rc,xu2019pc} and ProxylessNAS~\cite{han2019proxyless} formulate the search space as a super-graph where the probability to adopt an operator is represented by a continuous importance weight, allowing an efficient search of the architecture using gradient descent. 
%Based on it, works~\cite{jin2019rc,xu2019pc,,bender2018understanding,stamoulis2019single,stamoulis2019single,yu2020bignas,mei2020atomnas} model NAS as a single training process of an over-parameterized network that comprises all candidate paths.
%Another set of works~\cite{you2020greedynas,guo2019single,li2020random,chu2019fairnas,xie2018snas,chu2019scarletnas} directly search the discrete search space by sampling and optimizing a single path in a supernet.
Other approaches~\cite{bender2018understanding,stamoulis2019single,guo2019single} utilize a random sampling approach when training the super-net and search for the best model candidate after the network converges.
Inspired by the manually designed structures, ~\cite{tan2019mnasnet,howard2019searching} use a search space based on MobileNetV2~\cite{sandler2018mobilenetv2} to search for efficient structures. Mixed convolution~\cite{tan2019mixconv,mei2020atomnas} is also adopted in NAS search spaces due to its multi-scale feature modeling capability.
Recently, model scaling techniques are used to expand the search space from operators to other hyper-parameters such as input resolutions, channel numbers, and layer numbers~\cite{cai2019once,yu2020bignas}.
In order to search for efficient models, the existing methods usually borrow efficient operators from manually designed networks, such as depthwise convolution and Inverted Residual Block~\cite{sandler2018mobilenetv2}. To construct the search space with more powerful operators, we design a new efficient Transformer structure that can be directly inserted into existing NAS search spaces.

\noindent\textbf{Neural Architecture Search for Dense Prediction.}
%Most existing NAS methods focus on searching either stacked operators or repeated cell-structured directed acyclic graph (DAG) for the classification task.
The current NAS algorithms either reuse search spaces for image classification or only search for a feature aggregation head for dense prediction tasks. A single branch super-net structure is usually utilized for dense prediction tasks such as semantic segmentation~\cite{shaw2019squeezenas,liu2019auto,lin2020graph}, object detection~\cite{du2020spinenet,chen2019detnas,ghiasi2019fpn}, and human pose estimation~\cite{dai2020data}. Structures of feature aggregation head are also discovered using NAS algorithms for semantic segmentation~\cite{chen2018searching,nekrasov2019fast}.
%DA-NAS~\cite{dai2020data} searches with an operator combination of inverted residual blocks, shuffle blocks and residual blocks for image classification and human pose estimation. Similar search space and strategy are also explored in object detection~\cite{chen2019detnas,ghiasi2019fpn,du2020spinenet,wang2020fcos} and semantic segmentation~\cite{shaw2019squeezenas,liu2019auto,nekrasov2019fast,lin2020graph,chen2018searching}.
%However, such methods organize network cells in a chain-like single branch manner and then introduce heavy heads (\eg, ASPP~\cite{chen2017deeplab}) for multi-scale information, making them inefficient and non-generalizable.
Recent explorations~\cite{gong2020autopose, zhang2020dcnas} aim to find an optimal network layout in a hierarchical multi-scale search space. However, their search spaces use fixed width of layers which result in computationally heavy models. In contrast, we propose a multi-branch search space where each branch specializes for a typical feature resolution. The same search space can be directly used for various dense prediction tasks that have different preferences on the granularity of features, in which the computation budget is allocated for different resolutions through an end-to-end optimization. 

\section{Methodology}
Our method aims to search for network structures within a multi-branch search space containing both Convolutions and Transformers with a resource-aware search strategy.
In this section, we first introduce our lightweight Transformers. We then detail our multi-branch search space and how to integrate our Transformers into it. Finally, we describe the resource-aware fine-grained search strategy.

\begin{figure}[t]
    \centering
    \includegraphics[width=0.98\linewidth]{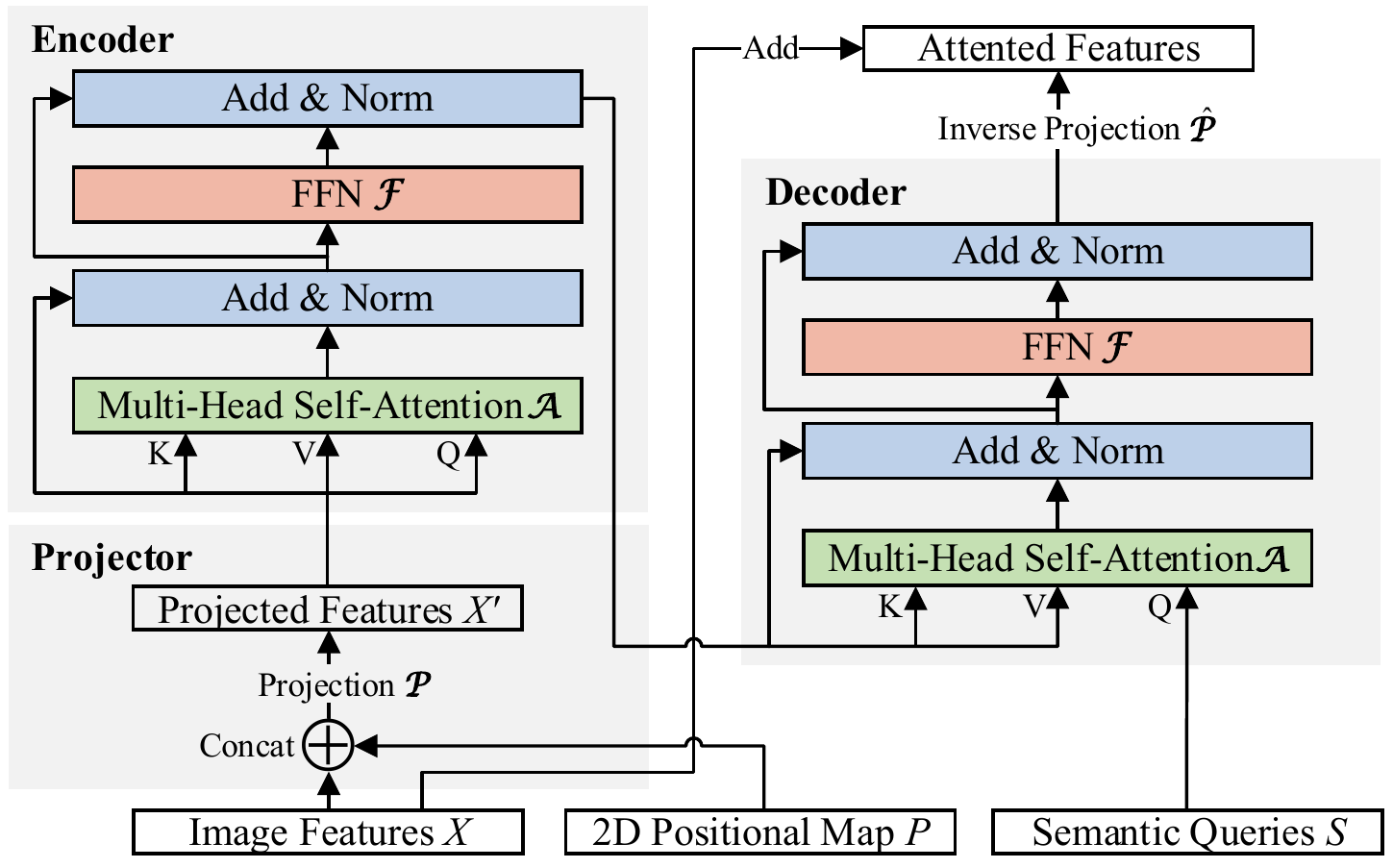}
    \caption{The architecture of our lightweight Transformer, which contains a projector, an encoder, and a decoder. It can be used plug-and-play to enhance the global context of image features.
    % See Sec.~\ref{sec:tranformer} for details.
    }
    \label{fig:transformer}
    \vspace{-4pt}
\end{figure}

\begin{figure*}[t]
    \centering
    \includegraphics[width=0.96\textwidth]{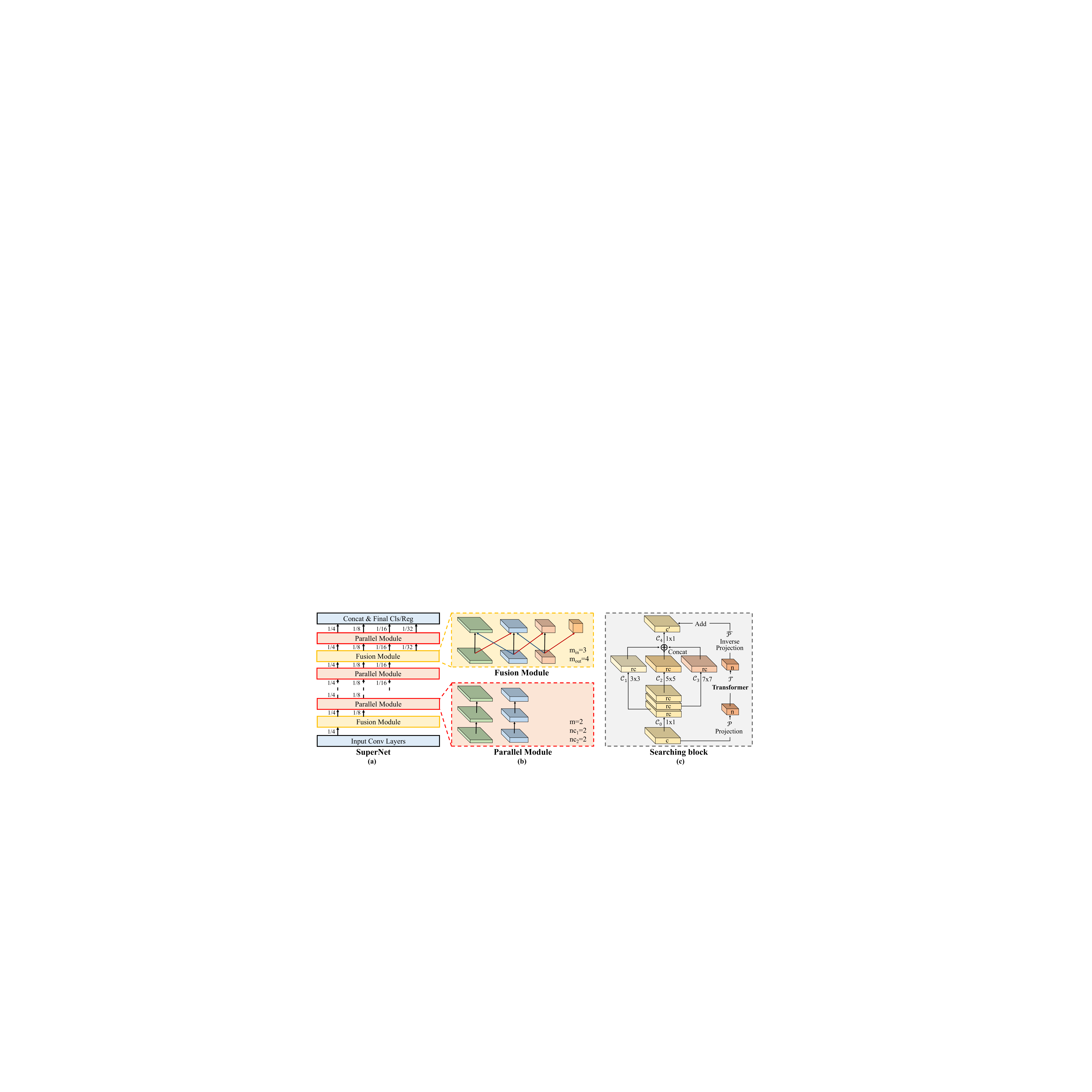}
    \caption{
    (a) Our multi-branch search space, which is composed of parallel modules and fusion modules alternately.
    ``$1/4, 1/8, \ldots$" denote the down-sampling ratios.
    (b) Illustration of parallel modules and fusion modules. The red, black, and blue arrows represent the reduction \cells, normal \cells, and normal \cells~with upsampling, respectively. The cubes represent feature maps. In this example, the fusion module generates an extra branch from the previous lowest-resolution branch by a reduction \cell.
    (c) The proposed \cell~that contains both convolutions with different kernel sizes $\mathcal{C}_1,\mathcal{C}_2,\mathcal{C}_3$ and a lightweight Transformer $\mathcal{T}$.
    % See Sec.~\ref{sec:space} for details.
    }
    \label{fig:space}
    \vspace{-6pt}
\end{figure*}

\subsection{Lightweight Transformers}
\label{sec:tranformer}
The standard Transformer~\cite{vaswani2017attention,carion2020end} cannot be directly applied to high-resolution images and mobile scenarios, as its computational cost grows quadratically to the number of pixels. Our lightweight Transformer shown in Fig.~\ref{fig:transformer}, which consists of a projector, an encoder, and a decoder, is proposed to solve this issue (see Fig.~\ref{fig:transformer}).

\noindent\textbf{Projector.}~~To reduce the computational costs, we project the input feature $X\in\mathbb{R}^{c\times h \times w}$, together with the positional encoding, into a reduced size of $n\times s\times s$ by a projection function $\mathcal{P}(\cdot)$, where $n$ denotes the number of queries and $s\times s$ is the reduced spatial size.
Formally, the projection process can be represented by:
\begin{equation}
\footnotesize
X' = \mathcal{P}(\mathrm{Concat}(X, P)),
\label{eq:projector}
\end{equation}
where $\mathrm{Concat}$ denotes the concatenation operator, $P\in\mathbb{R}^{d_p\times h \times w}$ is a positional encoding which compensates for the loss of spatial information during the self-attention process, and $X'\in\mathbb{R}^{n\times s^2}$ is the projected and flattened embedding. The projector $\mathcal{P}$ first uses a point-wise convolution (with a Batch Normalization layer) to reduce the channel dimension of the feature map from $c+d_p$ to a smaller dimension $n$ and then uses a bilinear interpolation operation to resize the spatial dimension of the feature map to $s\times s$.
The positional encoding $P$ in Eq.~\ref{eq:projector}, is simply a normalized 2D positional map: 
\par\nobreak
\vspace{-8pt}
\begin{footnotesize}
\begin{align}
& P[0,i,j] = i/h,~~~~~i \in [0,h-1] \notag \\
& P[1,i,j] = j/w,~~~~~j \in [0,w-1]
\label{eq:position}
\end{align}
\end{footnotesize}
\hspace{-3pt}The 2D positional map $P$ is very efficient as it contains only 2 channels (\ie, $d_p=2$). Later in the experiments, we show that this simple encoding outperforms the sinusoidal positional encoding~\cite{vaswani2017attention} and the learned embedding~\cite{carion2020end}.

\noindent\textbf{Encoder.}~~After the projection, the original feature $X$ is transformed into a set of $n$ tokens $X'$; each token is an $s^2$-dimensional semantic embedding with positional information. $X'$, is then fed into our encoder as queries, keys, and values $Q,K,V\in\mathbb{R}^{n\times s^2}$. Following the standard Transformer~\cite{vaswani2017attention}, our lightweight Transformer is built upon the Multi-Head Self-Attention $\mathcal{A}(\cdot)$, which allows the model to jointly attend to information at different positions. It is defined as:%
\par\nobreak
\vspace{-8pt}
\begin{footnotesize}
\begin{align}
\mathcal{A}_{enc}(Q, K, V) & = \mathrm{Concat}(\mbox{head}_1,\ldots,\mbox{head}_h)W^O \notag \\
\text{where}~~\mbox{head}_i & = \mathrm{Attention}(QW_i^Q, KW_i^K, VW_i^V) \notag \\
& = \softmax \left[\frac{QW_i^Q(KW_i^K)^T}{\sqrt{d}}\right]VW_i^V
\label{eq:selfattention}
\end{align}
\end{footnotesize}
\hspace{-3pt}where $h$ is the number of heads, $d$ is the hidden dimensions of the attended subspaces, and
$W_i^Q, W_i^K, W_i^V\in\mathbb{R}^{s^2\times d}, W^O\in\mathbb{R}^{hd\times s^2}$ are learned embeddings (weights).
A position-wise Feed-Forward Network (FFN) $\mathcal{F}_{enc}(\cdot)$, which consists of two linear transformations with a ReLU activation in between, is then applied to the attended features:
\begin{equation}
\footnotesize
\mathcal{F}_{enc}(x) = \mathrm{max}(0, xW_1 + b_1)W_2 + b_2
\end{equation}
where $W_1\in\mathbb{R}^{s^2\times 4s^2}, W_2\in\mathbb{R}^{4s^2\times s^2}, b_1$ and $b_2$ are the weights and biases of the linear layers respectively.
%As shown in Fig.~\ref{fig:transformer}, the whole process of our Transformer encoder can be represented by $\mathcal{F}(\mathcal{A}(Q, K, V))$, where the token-wise attention $A\in\mathbb{R}^{n\times n}$ is first calculated, and linear transformations are then applied across all spatial-wise positions to get the global attended feature $F$.
We employ residual connections \cite{he2016deep} around both the Multi-Head Self-Attention layer and the Feed-Forward Network, which are followed by layer normalization~\cite{ba2016layer} as in ~\cite{vaswani2017attention}.

\noindent\textbf{Decoder.}~~Our decoder follows the same paradigm of the encoder: Multi-Head Self-Attention layer and Feed-Forward Networks. It uses the output of the encoder $F$ as keys and values and a set of n learnable $s^2$-dimensional semantic embeddings $S\in\mathbb{R}^{n\times s^2}$ as queries. The decoder can be formalized as $\mathcal{F}_{dec}(\mathcal{A}_{dec}(S, F, F))$.
%It further enhances the self-attention and global context by leveraging a learned semantic embedding.
%
Finally, the output of the decoder is transformed back to the proper shape by an inverse projection function $\widehat{\mathcal{P}}(\cdot)$. Like the projector $P$, the inverse projector $\widehat{\mathcal{P}}(\cdot)$ consists of a point-wise convolution (with a Batch Normalization layer) and a bilinear interpolation operation.
Note that since image modeling is not a sequence prediction task, and there is no temporal relationship between the semantic tokens, we remove the first Multi-Head Attention in the standard Transformer decoder~\cite{vaswani2017attention}.

\label{sec:complexity}
\noindent\textbf{Time Complexity.}~~The time complexities of our Multi-Head Self-Attention and our FFN are $O(4nds^2 + 2n^2d)$ and $O(8ns^4)$, respectively, where $s^2$, $d$ and $n$ are in the projected low-dimensional space. Since $s^2$ is a projected small spatial size, the overall time complexity (FLOPs) $O_{\mathcal{T}}(n)$ of our Transformer is approximately linear with $n^2d$. In the following part, we will further introduce a fine-grained search strategy to reduce the number of tokens $n$ in order to make the lightweight Transformer more efficient.

In summary, the \textbf{main difference} between our lightweight Transformer and the standard Transformer~\cite{vaswani2017attention,carion2020end} lies in:
(1) A projection function $\mathcal{P}(\cdot)$ is used to learn self-attention in a low-dimensional space.
(2) A simpler yet effective 2D positional map $P$ is used for positional encoding.
(3) The first Multi-Head Attention and the spatial encoding in the standard Transformer decoder are removed.
%(4) The output of our encoder is directly used as the key and the value of the decoder with residual connections. However, the residual path is connected to the query in~\cite{vaswani2017attention}.

\subsection{Multi-branch Search Space}
\label{sec:space}
Inspired by HRNet~\cite{wang2020deep}, we design a multi-branch search space for dense predictions that contains both multi-scale features and global contexts while maintaining high-resolution representations throughout the network. 
% In the following, we first outline the entire network architecture.
% We then detail our \cell, which is the key component of the search space containing both Convolutions and Transformers. 

\noindent\textbf{Overview.}~~The network consists of two modules: the parallel module and the fusion module. Both of the two modules are constructed with our \cells. As shown in Fig.~\ref{fig:space} (a), after two convolutions which decrease the feature resolution to $1/4$ of the input image size, we start with this high-resolution branch and gradually add high-to-low resolution branches through fusion modules, and connect the multi-resolution branches in parallel through parallel modules. Finally, multi-branch features are resized and concatenated together, and connected to the final classification/regression layer without any additional heads.

The parallel module obtains larger receptive fields and multi-scale features by stacking \cells~in each branch. It has $m\in[1,4]$ branches containing $nc_1,\ldots,nc_m$ convolutions with $nw_1,\ldots,nw_m$ channels in each branch.
%Formally, a parallel module can be represented as $[m, [nc_1,\ldots,nc_m], [nw_1,\ldots,nw_m]]$ (see Fig.~\ref{fig:space} (b)).
%
A fusion module is used after a parallel module to exchange information across multiple branches. An extra lower-resolution branch is also generated from the previously lowest-resolution branch until it reaches $1/32$ downsampling ratios. A fusion module takes $m_{in}$ branches from the previous parallel module as input and outputs $m_{out}$ branches.
%between two parallel modules with $m_{in}$ and $m_{out}$ branches to perform feature interaction between multiple branches by element-wise addition. 
For each output branch, all its neighboring input branches are fused by using the \cell~to unify their feature map sizes. For example, a $1/8$ output branch integrates information of $1/4, 1/8, \mbox{and}~1/16$ input branches. In our fusion module, the high-to-low resolution feature transformation is realized by the reduction \cell, while the low-to-high resolution feature transformation is implemented with the normal \cell~and upsampling.  

\noindent\textbf{\Cell.}~~As shown in Fig.~\ref{fig:space} (c), our \cell~contains two paths: one path is a MixConv~\cite{tan2019mixconv}, the other path is a lightweight Transformer which aims to provide more global contexts. The number of convolutional channels and the number of tokens in the Transformer are searchable parameters. 

Formally, let $X$ be the input of $c$ feature channels (the spatial dimension is omitted for simplicity). In the MixConv path, the first layer is a point-wise convolution $\mathcal{C}_0$ which expands $X$ to a $3r\times c$ dimension (\ie, the expansion ratio is $3r$); the output is split into three parts with an equal number of channels (\ie, each with $r\times c$ channels), which are then fed into three depth-wise convolutions $\mathcal{C}_1,\mathcal{C}_2,\mathcal{C}_3$ with kernel sizes of $3\times 3, 5\times 5$, and $7\times7$, respectively.
The outputs of these three layers are concatenated, followed by another point-wise convolution $\mathcal{C}_4$ that produces the feature map with the desired number of channels $c'$.
In the Transformer path, a lightweight Transformer $\mathcal{T}$ with $n$ tokens is applied to the input feature $X$ to obtain the global self-attention. The outputs of two branches are added as the final output of the \cell. Intuitively, the Transformer path can be regarded as a residual path for enhancing the global context within the \cell.
The information flow in a \cell~can be written as:
\begin{equation}
\footnotesize
X' = \mathcal{C}_4(\mathrm{Concat}(
\mathcal{C}_1(\mathcal{C}_0(X)_1),
\mathcal{C}_2(\mathcal{C}_0(X)_2),
\mathcal{C}_3(\mathcal{C}_0(X)_3))) + \mathcal{T}(X)
\end{equation}
where $\mathcal{C}_0(X)_i$ represents the $i$-th part of the output of $\mathcal{C}_0(X)$, as shown in Fig.~\ref{fig:space} (c).
Note that when the strides of the convolutions $\mathcal{C}_1, \mathcal{C}_2, \mathcal{C}_3$ are equal to $2$, as in the reduction \cell,~the inverse projection $\widehat{\mathcal{P}}(\cdot)$ in Transformer resizes its input into half size of the original spatial dimension in order to match the output shape of $\mathcal{C}_4$. 
%In this way, the whole supernet is constructed by our normal and reduction \cells, making it easily fit for a limited computational budget by shrinking the depth-wise convolutional channels of $\mathcal{C}_1, \mathcal{C}_2, \mathcal{C}_3$ and quires/tokens of Transformer $\mathcal{T}$ while maintaining multi-scale and global information.

% \noindent\textbf{Advantages.}~~Our multi-branch network space can extract both multi-scale features and global contexts with Convolutions and Transformers. Convolutions with different kernel sizes have different reception fields. In addition, the Transformer has a global reception field. The multi-branch multi-scale search space with learnable reception fields can be customized for various tasks naturally.
% Moreover, each \cell~contains $3rc+n$ searchable sub-layers, which further makes fine-grained search strategies for mobile scenarios possible.
% The channels in depth-wise convolutions $\mathcal{C}_1, \mathcal{C}_2, \mathcal{C}_3$ and the queries/tokens in Transformers $\mathcal{T}$ can be removed partly or completely according to computing resources and tasks without affecting the overall network structure.
% To the best of our knowledge, we are the first to integrate Transformers with a searchable number of queries into the NAS search space in computer vision.

\subsection{Resource-aware Fine-grained Search}

Our supernet is a multi-branch network where each branch is a chain of \cells~operating at different resolutions; each \cell~combines a MixConv and a Transformer.
Unlike previous searching methods that are designed for specific tasks, we aim to customize the network for various tasks. Specifically, we propose a resource-aware channel/query-wise fine-grained search strategy to explore the optimal feature combination for different tasks.

We adopt a progressive shrinking NAS paradigm which generates lightweight models by discarding some of the convolutional channels and Transformer queries during training. 
As described in~\cite{mei2020atomnas}, as channels in depth-wise convolutions are independent in our \cell, any channels from these convolutions can be easily removed without affecting the other \cells; we only need to remove the corresponding weights from the convolutions.
%
%Similarly, with the projection $\mathcal{P}(\cdot)$ and the inverse projection $\widehat{\mathcal{P}}(\cdot)$, our Transformer $\mathcal{T}$ is designed to have a variable number of queries and tokens. If a query is discarded, the projections $\mathcal{P}(\cdot)$ and $\widehat{\mathcal{P}}(\cdot)$ will process $(n-1)\times s\times s$ sized features in the low-dimensional space. In this way, all tokens and features of both the Transformer encoder and decoder are automatically scaled.
Similarly, if a token of the Transformer is discarded, we just remove the corresponding weights from the $1\times1$ convolutions of the projections $\mathcal{P}(\cdot)$ and $\widehat{\mathcal{P}}(\cdot)$, and the corresponding embedding from queries $S$.

In the rest of this paper, we call a channel of the depth-wise convolutions or a token in Transformers a {\it search unit}. A \cell~with $c$ input channels, the expansion ratio of $3r$, and $n$ tokens has $3rc + n$ search units in total.

Following Darts~\cite{hanxiao2019darts}, we introduce an importance factor $\alpha>0$ that can be learned jointly with the network weights for each search unit of the \cell. We then progressively discard those with low importance while maintaining overall performance.
Inspired by works on channel pruning~\cite{yu2018slimmable,liu2017learning,mei2020atomnas}, we add a resource-aware L1 penalty on $\alpha$, which effectively pushes importance factors of high computational costs to zero.
Specifically, the L1 penalty of a search unit is weighted by the amount of the reduction in computational cost $\Delta>0$ (\ie FLOPs in this case):
\begin{align}
\footnotesize
\Delta_i = 
\begin{cases}
3 \times 3 \times h \times w, & i \in [0,rc) \\
5 \times 5 \times h \times w, & i \in [rc,2rc) \\
7 \times 7 \times h \times w, & i \in [2rc,3rc) \\
O_{\mathcal{T}}(n') - O_{\mathcal{T}}(n'-1), & i\in [3rc,3rc+n) \\
\end{cases}
\label{eq:penalty}
\end{align}
where $O_{\mathcal{T}}$ is the FLOPs of the Transformer defined in Sec.~\ref{sec:complexity}, $i$ is the index of the search unit, $n'$ is the number of remaining tokens. Note that $\Delta$'s for search units of convolutions are fixed, while in the Transformer, $\Delta$'s is a function of the number of remaining tokens. It is worth mentioning that, although FLOPs is not always a good measure of latency, we use it anyway as it is the most widely and easily used metric. The Eq.~\ref{eq:loss} can be easily adapted to use other metrics, \eg, latency and energy cost.

With the added resource-aware penalty term, the overall training loss is:
\begin{equation}
\footnotesize
    L = L_\text{task} + \lambda \sum_{i\in[0,3rc+n)} \Delta_i |\alpha_i|
\label{eq:loss}
\end{equation}
where $L_\text{task}$ denotes the standard classification/regression loss, and $\lambda$ denotes the coefficient of the L1 penalty term. 
%The weight decay can help constrain the value of the network weight to prevent it from being too large and making importance factors $\alpha$ difficult to learn.

During training, after every few epochs, we progressively remove the search units whose importance factors are below a predefined threshold $\epsilon$ and re-calibrate the running statistics of Batch Normalization (BN) layers. 
Note that if all tokens of a Transformer are removed, the Transformer will degenerate into a residual path, as shown in Fig.~\ref{fig:transformer}.

When the search ends, the remaining structure not only represents the best accuracy-efficiency trade-offs, but also has the optimal low-level/high-level and local/global feature combination for a specific task. In addition, since the network training and architecture search are conducted in a unified end-to-end manner, the resulting network can be used directly without fine-tuning.

%With the help of the resource-aware L1 regularization, our method is able to find the best accuracy-efficiency trade-offs for different of resource budgets.  

%Moreover, our multi-branch supernet can be customized for different tasks during the search process.
%Different convolutional channels and Transformer tokens of different branches are retained for different tasks, \ie, our method can find the optimal low-level/high-level and local/global feature combination for a specifictask.
\vspace{-4pt}
\section{Experiments}

\vspace{-3pt}
\subsection{Implementation Details}
%versatility
To validate the generalizability of our method, we select five benchmark datasets on four representative tasks for performance evaluation: image classification on ImageNet~\cite{deng2009imagenet}, human pose estimation on COCO keypoint~\cite{lin2014microsoft}, semantic segmentation on Cityscapes~\cite{cordts2016cityscapes} and ADE20K~\cite{zhou2017scene}, and 3D object detection on KITTI~\cite{geiger2012we}. These benchmarks are carefully selected as they require different receptive fields, global/local contexts, and 2D/3D perceptions.
In this work, the same supernet is used for all five benchmarks; It begins with two $3\times 3$ convolutions with stride 2, which is followed by five parallel modules (respectively with 1, 2, 3, 4, 4 branches); a fusion module is inserted between every two adjacent parallel modules, to obtain multi-scale features. %The details of the architecture are provided in Supplemental Materials.
For Transformers, we set $s=8$, $d=s^2=64$, and $h=1$. In some evaluation experiments without search, we fix $d=8$.
The expansion ratio $r$ of the \cell~is set to be 4. For the MixConv, we use the scales from the batch normalization layers after the depth-wise convolutions as the importance factors; for the Transformer, we use the scales from the batch normalization layer in the projector $\mathcal{P}$ as the importance factors. On each benchmark, we obtain \alias-A and \alias-B using different $\lambda$ values. Search units with $\alpha<0.001$ are deemed unimportant and removed every five epochs. 
% All the models are trained from scratch on 8 Tesla V100 GPUs.
Unless specified, all experiments in this paper use standard training protocols, \eg, we don't apply techniques like AutoAug~\cite{cubuk2018autoaugment}, Mixup~\cite{zhang2017mixup}, and Cutout~\cite{devries2017improved}. All our models are trained from scratch without pretraining on the ImageNet dataset, and are evaluated with single-scale input and without multi-crop.
Details of the datasets and the training settings for each task can be found in Supplemental Materials.

\begin{table}[t]
\caption{Comparision with state-of-the-arts on ImageNet under the mobile setting.
$^\dag$ denotes methods using Swish activation~\cite{ramachandran2017searching}, $^\ddag$ denotes methods trained on AutoAugment~\cite{cubuk2018autoaugment} or RandAugment~\cite{cubuk2020randaugment}.
FLOPs is measured using an input size of $224\times 224$.
}
\centering
\scalebox{1}{
\footnotesize
\tabcolsep=0.26cm
\begin{tabular}{lrrc}
\toprule
Model & Params & FLOPs & Top-1(\%) \\%mingyu & Top-5(\%) \\
\midrule
CondenseNet \cite{huang2018condensenet}  & 2.9M & 274M & 71.0 \\%mingyu & 90.0 \\
% CondenseNet (G=C=4)  & 4.8M & 529M & 73.8 & 91.7 \\
ShuffleNetV1 \cite{zhang2018shufflenet}          & 3.4M & 292M & 71.5 \\%mingyu& - \\
ShuffleNetV2 \cite{ma2018shufflenet}          & 3.5M & 299M & 72.6 \\%mingyu& - \\
% ShuffleNetV2 2$\times$      & 7.4M & 591M & 74.9 & - \\
% MobileNetV1 \cite{howard2017mobilenets}           & 4.2M & 575M & 70.6 \\%mingyu& 89.5 \\
MobileNetV2 \cite{sandler2018mobilenetv2}        & 3.4M & 300M & 72.0 \\%mingyu& 91.0 \\
MobileNetV3 \cite{howard2019searching}$^\dag$ & 5.4M & 219M & 75.2 \\%mingyu& - \\
% ResNet50 \cite{he2016deep} & 25.0M & 4.12G & 75.3 \\
EfficientNet-B0 \cite{mingxing2019efficient}$^{\dag\ddag}$ & 5.3M & 390M & 77.3 \\%mingyu& 93.5 \\
\midrule
% DARTS \cite{hanxiao2019darts}      & 4.9M & 595M & 73.1 \\%mingyu& - \\
% NASNet-A \cite{zoph2017nasnet}     & 5.3M & 564M & 74.0 \\%mingyu& 91.6 \\
% FBNet-A \cite{wu2019fbnet}         & 4.3M & 249M & 73.0 & - \\
FBNet-B \cite{wu2019fbnet}         & 4.5M & 295M & 74.1 \\%mingyu& - \\
% FBNet-C \cite{wu2019fbnet}         & 5.5M & 375M & 74.9 & - \\
AutoSlim-MobileNetV2 \cite{yu2019autoslim} & 5.7M & 305M & 74.2 \\%mingyu& - \\
Proxyless \cite{han2019proxyless}                  & 4.1M & 320M & 74.6 \\%mingyu& 92.2 \\
% Proxyless (GPU)                   & 7.1M & 465M & 75.1 & 92.5 \\
DA-NAS \cite{dai2020data} & $-$ & 323M & 74.3 \\
AtomNAS-A \cite{mei2020atomnas} & 3.9M & 258M & 74.6 \\%mingyu& 92.1 \\
% AtomNAS-C \cite{mei2020atomnas} & 4.7M & 360M & 75.9 \\%mingyu& 92.7 \\
SinglePathOneShot \cite{guo2019single} & 3.4M & 328M & 74.7 \\%mingyu& 92.0 \\
FairNAS-C \cite{chu2019fairnas} & 4.4M & 321M & 74.7 \\
% PDARTS \cite{chen2019progressive}       & 4.9M & 557M & 75.6 & 92.6 \\
% DenseNAS-A \cite{fang2020densely} & 7.9M & 501M & 75.9 & 92.6 \\
% DARTS+ \cite{liang2019darts+} & 5.1M & 591M & 76.3 & 92.8 \\
MnasNet-A1 \cite{tan2019mnasnet}      & 3.9M & 312M & 75.2 \\%mingyu& 92.5 \\
% MnasNet-A2$^\dag$ \cite{tan2019mnasnet}     & 4.8M & 340M & 75.6 & 92.7 \\
TF-NAS-C \cite{hu2020tf} & $-$ & 284M & 75.2 \\
% AtomNAS-B \cite{mei2020atomnas} & 4.4M & 326M & 75.5 \\%mingyu& 92.6 \\
% SCARLET-C \cite{chu2019scarletnas} & 6.0M & 280M & 75.6 \\%mingyu& 92.6 \\
SCARLET-B \cite{chu2019scarletnas} & 6.5M & 329M & 76.3 \\
ST-NAS-A \cite{guo2020powering} & 5.2M & 326M & 76.4 \\
\midrule
\textbf{\alias-A} & 5.5M & 267M & 75.7 \\%mingyu& 92.3 \\
\textbf{\alias-B} & 6.4M & 325M & \textbf{76.5} \\

\midrule
MixNet-S \cite{tan2019mixconv}$^\dag$      & 4.1M & 256M & 75.8 \\%mingyu& 92.8 \\
% MixNet-M$^\dag$ \cite{tan2019mixconv}     & 5.0M & 360M & 77.0 & 93.3 \\
% SE-DARTS+$^{\dag \ddag}$ \cite{liang2019darts+} & 6.1M & 594M & 77.5 & 93.6 \\
AtomNAS-A+ \cite{mei2020atomnas}$^\dag$ & 4.7M & 260M & 76.3 \\%mingyu& 93.0 \\
% AtomNAS-C+ $^\dag$ \cite{mei2020atomnas}& 5.9M & 363M & 77.6 \\%mingyu& 93.6 \\
Once-for-all \cite{cai2019once}$^\dag$ & 4.4M & 230M & 76.0 \\%mingyu& - \\
Once-for-all (finetuned) \cite{cai2019once}$^\dag$ & 4.4M & 230M & 76.9 \\%mingyu& - \\
BigNAS \cite{yu2020bignas}$^{\dag\ddag}$ & 4.5M & 242M & 76.5 \\%mingyu& - \\
FairNAS-C+ \cite{chu2019fairnas}$^\dag$ & 5.6M & 325M & 76.7 \\
% AtomNAS-B+ $^\dag$ \cite{mei2020atomnas}& 5.5M & 329M & 77.2 \\%mingyu& 93.5 \\
\midrule
\textbf{\alias-A} $^{\dag\ddag}$ & 5.5M & 267M & 76.6 \\%mingyu& - \\
\textbf{\alias-B} $^{\dag\ddag}$ & 6.4M & 325M & \textbf{77.3} \\
\bottomrule
\end{tabular}
}
\label{tab:overall_cls}
\vspace{-4pt}
\end{table}

\subsection{Comparative Results}
We conduct experiments against the state-of-the-art methods on five benchmarks: image classification on ImageNet (Tab.~\ref{tab:overall_cls}), semantic segmentation on Cityscapes (Tab.~\ref{tab:overall_cityscapes}), semantic segmentation on ADE20K (Tab.~\ref{tab:overall_ade20k}), human pose estimation on COCO keypoint (Tab.~\ref{tab:overall_keypoint}), and 3d object detection on KITTI (Tab.~\ref{tab:overall_3ddet}). From these tables we can see that:
(1) Our method achieves state-of-the-art performance on all three dense prediction tasks and competitive results on the classification task. Compared with other tasks, classification usually benefits less from multi-scale and global contexts because it aggregates position-invariant features through global pooling.
(2) Many existing methods, such as~\cite{yu2020bignas,lin2020graph,li2019dfanet} utilize additional modules or pretraining on the ImageNet dataset to get better performance for a specific task. In contrast, our method is able to show superior results across multiple challenging datasets without any bells and whistles.
(3) We evaluate the mean and standard deviation of 5 runs on Cityscapes~\cite{cordts2016cityscapes} with Random Search~\cite{li2020random} as a baseline. It shows that our method yields stable results with a standard deviation of about only 0.3.
(4) For NAS methods toward high segmentation accuracy~\cite{liu2019auto,du2020spinenet} rather than accuracy-efficiency trade-offs, we reduce their network width to 1/2 (and depth to 1/2 for~\cite{liu2019auto}), thus obtain the tiny variants. Our method outperforms the second-best competitor by a large margin on Cityscapes (74.18 vs. 76.01), ADE20K (33.41 vs. 34.92), and COCO keypoint (74.9 vs. 75.5) using a much lighter model, showing its superiority and accuracy-efficiency balance ability on dense prediction tasks.

\begin{table}
  \caption{Comparative results on the CityScapes validation set (mIoU,\%).
  * indicates the model is pre-trained on the ImageNet dataset.
  FLOPs is measured using an input size of $512\times 1024$.
  $\dagger$ denotes the model is reduced by us for acc-efficiency trade-offs.
  }
  \centering
  \scalebox{0.95}{
  \tabcolsep=0.16cm
  \footnotesize
  \begin{tabular}{lrrc}
    \toprule        
    Model & Params & FLOPs & mIoU(\%) \\
    \midrule
    SegNet~\cite{badrinarayanan2017segnet} & 29.47M & 649G & 57.00 \\
    Enet~\cite{paszke2016enet} & 0.37M & 8.69G & 58.30 \\
    BiSeNet~\cite{yu2018bisenet} & 5.8M & 6.58G & 69.00 \\
    MobileNetV2~\cite{sandler2018mobilenetv2} & 2.11M & 5.33G  & 70.71 \\
    % MobileNetV3-Small~\cite{howard2019searching} & 0.47M & 0.74G & 68.38 \\
    MobileNetV3-Large~\cite{howard2019searching} & 1.51M & 2.48G & 72.36 \\
    HRNet-W$18$-Small~\cite{wang2020deep} & 3.94M & 19.30G & 75.44 \\
    \midrule
    C3~\cite{park2018c3} & 0.20M & 6.45G & 61.96 \\
    SkipNet-MobileNet~\cite{siam2018rtseg}* & $-$ & 13.80G & 62.40 \\
    EDANet~\cite{lo2019efficient}  & 0.68M & 7.98G & 65.11 \\
    SwiftNet~\cite{orsic2019defense} & 11.80M & 26G & 70.20 \\
    DFANet~\cite{li2019dfanet}* & 7.8M & 1.7G & 70.30 \\
    ShuffleNetV2+DPC~\cite{turkmen2019efficient} & 3.00M & 6.92G & 71.30 \\
    Auto-DeepLab~\cite{liu2019auto}-Tiny$\dagger$ & 3.16M & 27.29G & 71.21 \\
    GAS~\cite{lin2020graph}* & 1.50M & $-$ & 71.80 \\
    % SqueezeNAS-Small~\cite{shaw2019squeezenas} & 0.30M & 2.68G & 66.76 \\
    SqueezeNAS-Large~\cite{shaw2019squeezenas} & 0.73M & 8.35G & 72.40 \\
    % SqueezeNAS-XLarge~\cite{shaw2019squeezenas} & 1.80M & 19.43G & 74.54 \\
    % Auto-DeepLab-S [41] & 10.15 & 289.78 & 79.74 \\
    % SpineNet-49~\cite{du2020spinenet} & 21.86 & 145.87 & 77.34 \\
    SpineNet-49~\cite{du2020spinenet}-Tiny$\dagger$ & 5.49M & 37.99G & 74.18\\
    Random Search~\cite{li2020random} & 6.11$\pm$2.25M & 7.09$\pm$1.88G & 70.20$\pm$3.01 \\
    \midrule
    \textbf{\alias-A} & 2.20$\pm$0.14M & 1.91$\pm$0.11G & 74.26$\pm$0.37 \\%74.55  \\
    \textbf{\alias-B} & 3.85$\pm$0.19M  & 4.66$\pm$0.17G & \textbf{75.90}$\pm$0.30 \\%\textbf{76.01}  \\
    \bottomrule
  \end{tabular}
  }
  \label{tab:overall_cityscapes}
 \vspace{-3pt}
\end{table}

\begin{table}
  \caption{Comparative results on the ADE20K validation set (mIoU,\%).
  FLOPs is measured using an input size of $512\times 512$.}
  \centering
  \footnotesize
  \scalebox{1}{
  \tabcolsep=0.32cm
  \begin{tabular}{lrrc}
    \toprule        
    Model & Params & FLOPs & mIoU(\%) \\
    \midrule
    MobileNetV2\cite{sandler2018mobilenetv2} & 2.20M & 2.76G & 32.04 \\
    MobileNetV3-Large\cite{howard2019searching} & 1.60M & 1.32G & 32.31 \\
    HRNet-W$18$-Small~\cite{wang2020deep} & 3.97M & 10.23G & 33.41 \\
    \midrule
    \textbf{\alias-A} & 2.49M & 1.42G & 33.22 \\
    \textbf{\alias-B} & 3.86M & 2.19G & \textbf{34.92} \\
    \bottomrule
  \end{tabular}
  }
  \label{tab:overall_ade20k}
  \vspace{-8pt}
\end{table}

\vspace{-3pt}
\subsection{Ablation Study}
\vspace{-2pt}

\noindent\textbf{Search Space.}~~In this part, we study the design components of our search space. In Tab.~\ref{tab:ablation} we show how the introduction of different components affects the performance and FLOPs, using the Cityscapes segmentation benchmark as an example. The baseline search space, ``Single-branch" in Tab.~\ref{tab:ablation}, is a single-branch network with only $3 \times 3$ convolutions, where the up-sampling operations are applied at the end for dense prediction tasks. Adding the multi-branch architecture increases the mIoU from $66.23\%$ to $68.65\%$ with fewer parameters and FLOPs, showing the effectiveness of our multi-branch design. The MixConv with a mix of $3 \times 3, 5\times 5, 7\times 7$ convolutions in the \cell~further improves the mIoU by $3.34\%$. Finally, the lightweight Transformer provides another gain of $2.56\%$ ($71.99\%$ v.s. $74.55\%$) with only extra $70$M FLOPs.

\begin{table}[t]
\footnotesize
\centering
\caption{Comparisons on COCO keypoint validation set.
* indicates the model is pre-trained on the ImageNet dataset.
Params and FLOPs are calculated for the pose estimation network, and those for human detection and keypoint grouping are not included.}
\scalebox{0.85}{
\tabcolsep=0.08cm
\begin{tabular}{l|c|r|r|cccccc}
\toprule
Method & Input size & Params & FLOPs & 
$\operatorname{\mathbf{AP}}$ & $\operatorname{AP}^{M}$ & $\operatorname{AP}^{L}$ & $\operatorname{AR}$ \\
\midrule
ShuffleNetV1~\cite{zhang2018shufflenet}* & $256 \times 192$ & $1.0$M & $0.16$G & $58.5$ & $55.2$ & $64.6$ & $65.1$\\
ShuffleNetV2~\cite{ma2018shufflenet}* & $256 \times 192$ & $1.3$M & $0.17$G& $59.8$ & $56.5$ & $66.2$ & $66.4$\\
MobileNetV2~\cite{sandler2018mobilenetv2}* & $256 \times 192$ & $2.3$M & $0.33$G& $64.6$ & $61.0$& $71.1$& $70.7$ \\
NAS-CSS~\cite{nekrasov2019fast} & $256\times 192$&  $2.9$M & $1.48$G &
$65.9$&$63.1$&$70.0$&$69.3$  \\
% DeepLab v3+~\cite{chen2018encoder} & $256 \times 192$ & $5.8$M & $-$ &
% $66.8$ & $64.1$ & $70.7$ & $70.0$\\
DA-NAS~\cite{dai2020data} & $256\times 192$&  $10.9$M & $2.18$G & $68.4$ & $65.5$ & $74.4$ & $75.7$ \\
CPN~\cite{chen2018cascaded} & $256 \times 192$ & $27.0$M & $6.20$G &
$69.4$&$-$&$-$&$-$\\ 
SimpleBaseline-50~\cite{xiao2018simple}* & $256\times192$ &$34.0$M & $8.90$G  &${70.4}$ & ${67.1}$&${77.2}$&${76.3}$\\
HRNet-W$32$~\cite{wang2020deep} & $256\times 192$&  $28.5$M & $7.10$G &
$73.4$&$\mathbf{70.2}$&$80.1$&$78.9$  \\
AutoPose~\cite{gong2020autopose} & $256\times 192$&  $-$ & $10.65$G & $73.6$ & $69.8$ & $79.7$ & $78.1$ \\
\midrule
\textbf{\alias-A} & $256\times 192$ &  $1.7$M & $0.25$G & $67.7$&$65.4$&$71.1$&$70.8$ \\
\textbf{\alias-B} & $256\times 192$ & $6.1$M & $1.35$G & $\mathbf{73.7}$&$\mathbf{70.2}$&$\mathbf{80.6}$&$\mathbf{79.3}$ \\
\hline
\hline
ShuffleNetV1~\cite{zhang2018shufflenet}* & $384\times 288$ & $1.0$M & $0.35$G & $62.2$ & $57.8$ & $69.5$ & $68.4$\\
ShuffleNetV2~\cite{ma2018shufflenet}* & $384\times 288$ & $1.3$M & $0.37$G & $63.6$ & $59.5$ & $70.7$ & $69.7$\\
MobileNetV2~\cite{sandler2018mobilenetv2}* & $384\times 288$ & $2.3$M & $0.74$G & $67.3$ & $62.8$ & $74.7$ & $72.8$ \\
SimpleBaseline-50~\cite{xiao2018simple}* & $384\times 288$ &$34.0$M & $20.02$G &$72.2$ & $68.1$&$79.7$&$77.6$\\
% PoseNFS-1~\cite{yang2019pose} & $384\times 288$ &  $6.1$M & $4.0$G & $68.0$&$-$&$-$&$-$ \\
PoseNFS-3~\cite{yang2019pose} & $384\times 288$ &  $15.8$M & $14.8$G & $73.0$&$-$&$-$&$-$ \\
HRNet-W$32$~\cite{wang2020deep} &  $384\times 288$&  $28.5$M & 16.0G & $74.9$ & $71.5 $& $80.8$ &$79.3$  \\
\midrule
\textbf{\alias-A} & $384\times 288$ & 1.1M & 0.35G & $65.7$ & $62.5$ & $72.1$ & $71.4$ \\
% \textbf{\alias-A} & $384\times 288$ &  1.5M & 0.73G & $69.3$&$66.8$&$73.4$&$72.4$ \\
\textbf{\alias-B} & $384\times 288$ &  6.6M & 3.72G & $\mathbf{75.5}$&$\mathbf{72.6}$&$\mathbf{81.7}$&$\mathbf{79.4}$ \\
\bottomrule
\end{tabular}
}
\label{tab:overall_keypoint}
\vspace{-3pt}
\end{table}

\begin{table}[t]
    \caption{Vehicle 3D detection results(AP,\%) on the KITTI split1 validation set.
    All methods are implemented based on the Pointpillar~\cite{lang2019pointpillars} framework.
    FLOPs is calculated for 2D RPN network using an input size of $496\times 432$.}
    \centering
    {
    \footnotesize 
    \scalebox{1}{ 
    \tabcolsep=0.16cm
    \begin{tabular}{l|r|r|ccc}
        \toprule
        \multirow{1}{*}{Method} & Params & FLOPs & \textbf{Moderate} & Easy & Hard \\
        % & [split1/split2/test] & [split1/split2/test] & [split1/split2/test] \
        \midrule
	ShuffleNetV2~\cite{ma2018shufflenet} & 1.69M & 2.26G & 66.73 & 80.74 & 61.84 \\
	MobileNetV2~\cite{sandler2018mobilenetv2} & 2.49M & 6.47G & 67.65 & 82.52 & 64.22 \\
	Pointpillar~\cite{lang2019pointpillars} & 4.80M & 61.75G & 77.12 & 86.61 & 72.71 \\
	\midrule
	\textbf{\alias-A} & 2.13M & 3.22G & 69.74 & 83.09 & 64.89 \\
	\textbf{\alias-B} & 4.74M & 15.65G & \textbf{78.49}  & \textbf{87.62} & \textbf{75.53} \\
    \bottomrule
    \end{tabular}
    }}
    \label{tab:overall_3ddet}
    \vspace{-3pt}
\end{table}

\begin{table}[t]
    \caption{Ablation study of our search space on the CityScapes semantic segmentation validation set.}
    \centering
    {\footnotesize
    \scalebox{1}{ 
    \tabcolsep=0.1cm
    \begin{tabular}{l|r|r|ccc}
        \toprule
        \multirow{1}{*}{Method} & Params & FLOPs & \textbf{mIoU}(\%) & mACC(\%) & aACC(\%) \\
        \midrule
	Single-branch & 1.59M & 1.81G & 66.23 & 75.57 & 94.43 \\
	Multi-branch & 0.82M & 1.64G & 68.65 & 78.26 & 94.80 \\
	+MixConv & 1.12M & 1.86G & 71.99 & 80.33 & 95.40 \\
	+Transformer & 2.23M & 1.93G & \textbf{74.55} & \textbf{82.98} & \textbf{95.54} \\
    \bottomrule
    \end{tabular}
    }}
    \label{tab:ablation}
    \vspace{-10pt}
\end{table}

\begin{figure*}[t]
  \centering
    \begin{minipage}[t]{0.49\linewidth}
      \centering
      \includegraphics[width=\linewidth]{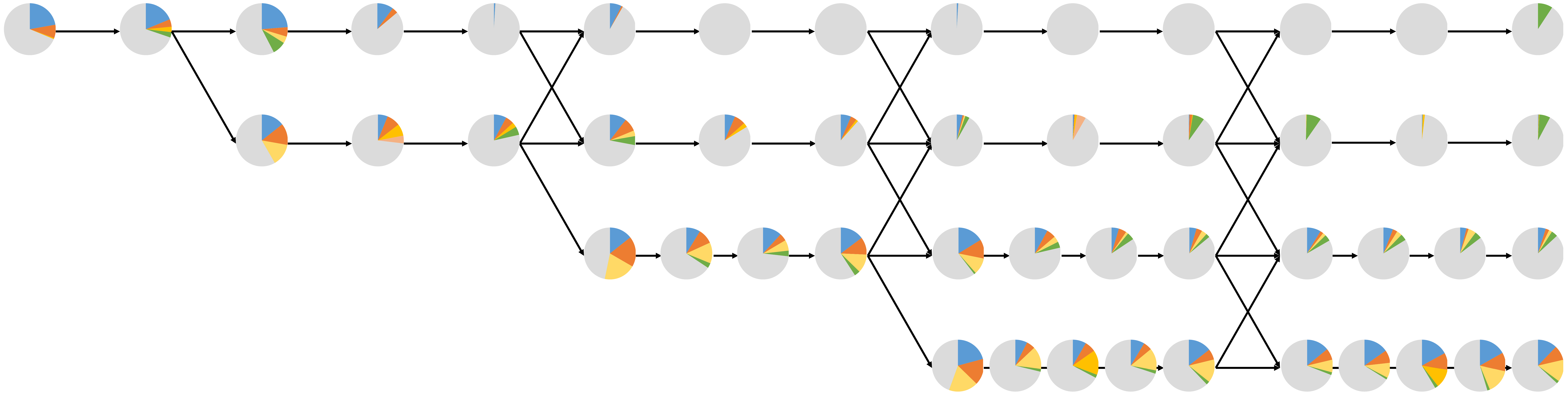}
      \small (a) Image classification on ImageNet 
    %   \vspace{6pt}
      \includegraphics[width=\linewidth]{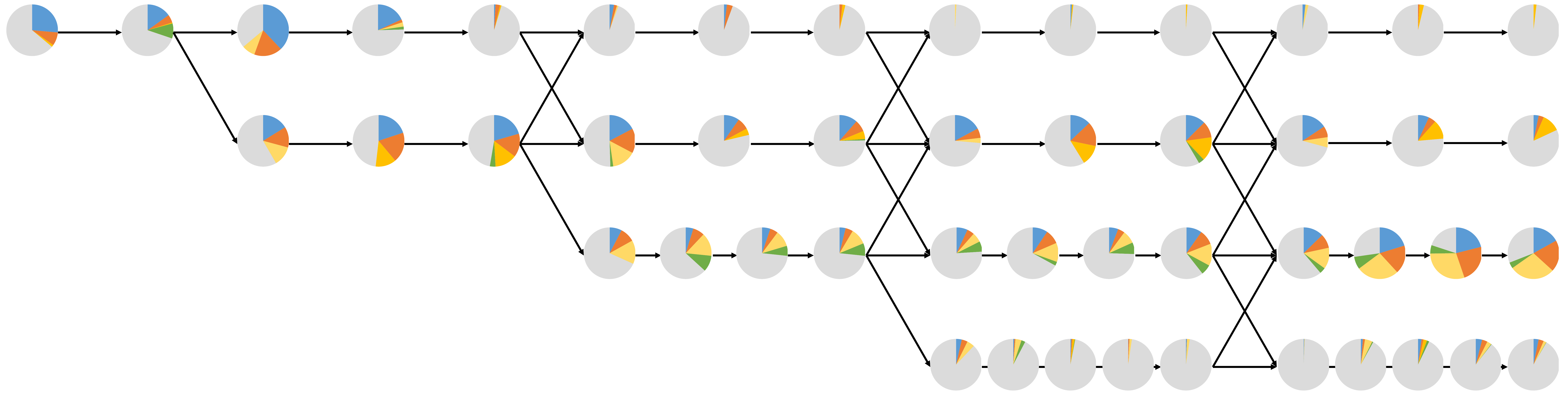}
      \small (c) Human pose estimation on COCO keypoint
      \vspace{4pt}
    \end{minipage}
  \hfill
    \begin{minipage}[t]{0.49\linewidth}
      \centering
      \includegraphics[width=\linewidth]{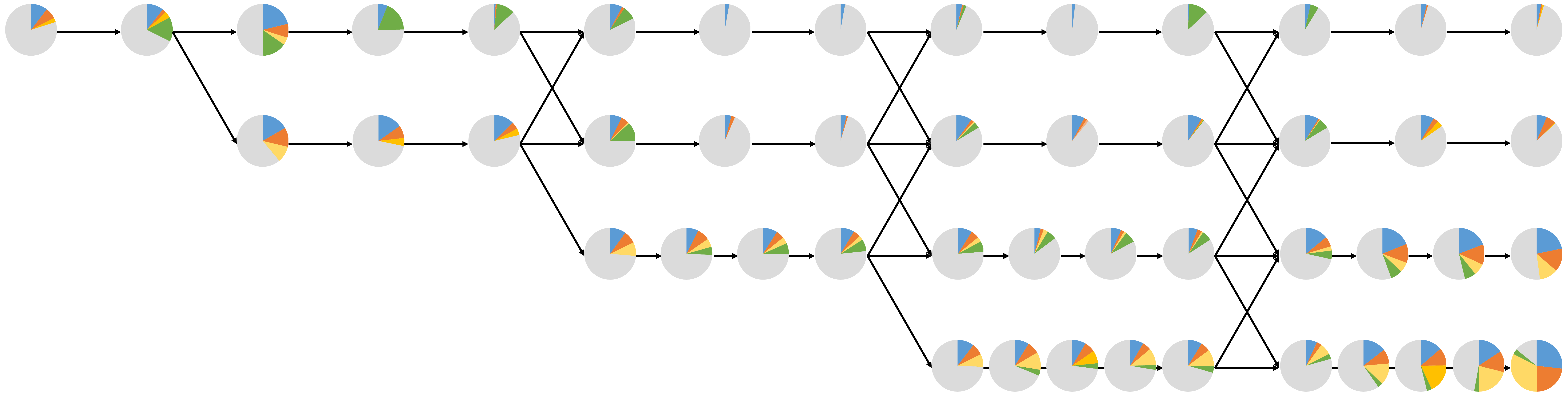}
      \small (b) Semantic segmentation on Cityscapes
    %   \vspace{6pt}
      \includegraphics[width=\linewidth]{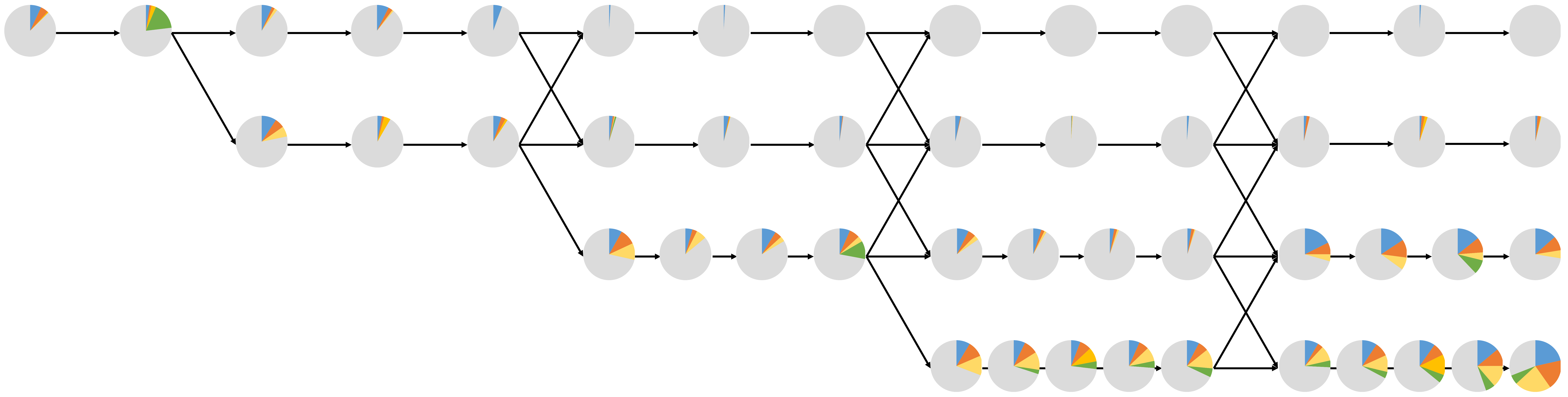}
      \small (d) 3D object detection on KITTI
      \vspace{4pt}
    \end{minipage}
  \caption{Visualization of the searched smaller architectures (\ie \alias-A) on four different tasks.
  The area of \textcolor{cyan}{\textbf{cyan}},
  \textcolor{red}{\textbf{red}},
  \textcolor{yellow}{\textbf{yellow}},
  \textcolor{green}{\textbf{green}},
  and \textcolor[RGB]{200,200,200}{\textbf{gray}} sectors indicate the number of $3\times3$, $5\times5$, $7\times7$ convolutional channels, the number of transformer queries, and the number of removed channels/queries, respectively.
  Note that if all queries and convolutional channels of a searching block are removed, the searching block will degenerate into a residual path.
  For simplicity, we only visualize searching blocks in the parallel module.
  We can see that our method is able to find different architectures for different tasks, showing that it can automatically adapt to various tasks. %Best viewd in color.
  }
  \label{fig:architecture}
  \vspace{-6pt}
\end{figure*}

\noindent\textbf{Lightweight Transformer.}~~In Tab.~\ref{tab:positional_encoding}, we study the choice of positional embeddings. It can be seen that using the proposed 2D positional map in the encoder of the Transformer achieves better performance than using the sinusoidal position encoding~\cite{vaswani2017attention} and the learned position embedding~\cite{devlin2018bert}. This may be because our lightweight Transformer has fewer queries and smaller token dimensions than the other two, and therefore it is unnecessary to use high dimension representation for position information. We also evaluate the alternative which uses the 2D positional map at both the encoder and the decoder; the performance is slightly worse than the encoder-only option.

The proposed Transformer can be used as a plug-and-play component. To show this, we add our Transformer to the Inverted Residual Blocks of two efficient models ShuffleNetV2~\cite{ma2018shufflenet} and MobileNetV2~\cite{sandler2018mobilenetv2}, and evaluate their performance on both ImageNet classification and Cityscapes segmentation tasks.
PSP module~\cite{zhao2017pyramid} is added as segmentation head to all models.
As shown in Tab.~\ref{tab:plug_and_play}, our Transformer improves the two baseline models on both classification and segmentation tasks. 
%This may be because dense prediction tasks require more global contexts.
%
% We then discuss two kinds of attention mechanisms by transposing the flattened featuremap: `channel' and `spatial' denote using each channel and spatial position of the featuremap as a token, respectively. As shown in Tab.~\ref{tab:transformer}, both the two lightweight transformers outperform their counterparts such as SE~\cite{hu2018squeeze} and Non-local~\cite{wang2018non} on dense prediction tasks. Since the channel-wise one shows better performance, we set it as default in this work.

% More ablative results can be found in Supplemental Materials.

\begin{table}[t]
    \caption{Comparisons (\%) of different positional embedding of Transformer with $n=8$ and $s=8$ on Cityscapes validation set. `Enc' and `Dec' denote the positional embedding are employed in the encoder and decoder of the Transformer, respectively.
    FLOPs is measured using an input size of $512\times 1024$.}
    \centering
    {\footnotesize
    \scalebox{0.98}{ 
    \tabcolsep=0.08cm
    \begin{tabular}{l|r|r|ccc}
        \toprule
        Input size & Params & FLOPs & \textbf{mIoU} & mACC & aACC \\
        \midrule
    Baseline & 1.120M & 1.863G & 71.99 & 80.33 & 95.40 \\
	Sinusoidal position encoding & 2.346M & 2.131G & 71.91 & 80.38 & 95.08 \\
	Learned position embedding & 2.930M & 2.236G & 72.71 & 80.88 & 95.33 \\
	2D positional map (Enc only) & 2.273M & 1.872G & \textbf{74.22} & \textbf{82.36} & \textbf{95.52}  \\
	2D positional map (Enc + Dec) & 2.278M & 1.873G & 73.70 & 81.89 & 95.47 \\
    \bottomrule
    \end{tabular}
    }}
    \label{tab:positional_encoding}
    \vspace{-4pt}
\end{table}

\subsection{Visualization of Searched Networks}
We visualized the four smaller models we found on each of the four benchmarks (\ie \alias-A) in Fig.~\ref{fig:architecture}. We can observe that our method can find different architectures for different tasks, showing that it can automatically adapt to various tasks:
(1) In the image classification task and the 3D detection task, at the high-resolution branches (\ie first and second branches), the models we found remove most of the search units; some searching blocks are even completely removed, as indicated by circles with complete gray in Fig.~\ref{fig:architecture}). The reason is that in these two tasks, global semantic information is more important than local information.
(2) The model for the segmentation task still retains computation from the first two branches, as it is important to keep high resolution imagery for semantic segmentation tasks.
(3) The human pose estimation model mainly utilizes the second and the third branches, which means it may rely more on middle-resolution semantics instead of high-resolution semantics.
(4) Transformers are more used in the segmentation and the human keypoint estimation tasks, indicating these dense prediction tasks benefit more from global contexts.

\begin{table}[t]
    \caption{Single-crop top-1 error rates (\%) on the ImageNet, and the mIoU (\%) on the cityscapes dataset. All models are trained from scratch. FLOPs is measured using classification models.}
    \centering
    {\footnotesize
    \scalebox{1}{ 
    \tabcolsep=0.13cm
    \begin{tabular}{l|rr|cc}
        \toprule
        Method & Params & FLOPs & top-1 & mIoU \\
    	\midrule
    	ShuffleNetV2~\cite{ma2018shufflenet} & 2.279M & 0.150G & 69.5 &  66.02 \\
    	ShuffleNetV2~\cite{ma2018shufflenet} + transformer & 2.758M & 0.157G & \textbf{70.1}  & \textbf{67.31} \\  
        \midrule
    	MobileNetV2~\cite{sandler2018mobilenetv2} & 3.505M & 0.319G & 72.0  & 68.98 \\
    	MobileNetV2~\cite{sandler2018mobilenetv2} + transformer & 3.770M & 0.321G & \textbf{72.8} & \textbf{70.17}\\
        \bottomrule
    \end{tabular}
    }}
    \label{tab:plug_and_play}
    \vspace{-10pt}
\end{table}

\section{Conclusion}
% In this paper, we introduce HR-NAS, which can find efficient and accurate networks for various computer vision tasks, by effectively encoding multiscale contextual information while maintaining high-resolution representations.
% In HR-NAS, we

% In this paper, we first introduce a lightweight and plug-and-play Transformer that can be easily combined with convolutional networks to enrich global semantic information for dense image prediction tasks.
% We effectively encoding both convolutions and the proposed Transformers into a well-designed high-resolution search space to model both multiscale contextual information and global semantic contexts.
% A channel/query-level fine-grained progressive shrinking strategy is then applied to the space for searching and customizing efficient models for various tasks.
% Our searched models achieve state-of-the-art trade-offs between performance and FLOPs for three dense prediction tasks and an image classification task, given small computational budgets.

In this paper, we introduce a lightweight and plug-and-play Transformer that can be easily combined with convolutional networks to enrich global contexts for dense image prediction tasks.
We then effectively encode both the proposed Transformers and convolutions into a well-designed high-resolution search space to model both global and multiscale contextual information.
A channel/query-level fine-grained progressive shrinking strategy is applied to the search space for searching and customizing efficient models for various tasks.
Our searched models achieve state-of-the-art trade-offs between performance and FLOPs for three dense prediction tasks and an image classification task, given only small computational budgets.

\noindent \textbf{Acknowledgements}~~Ping Luo was supported by the General Research Fund of HK No.27208720. Zhiwu Lu was supported by National Natural Science Foundation of China (61976220 and 61832017), and Beijing Outstanding Young Scientist Program (BJJWZYJH012019100020098).

%\appendix
\setcounter{section}{0}
\begin{appendices}

\section{Datasets and Settings}
In this section, we provide details of the datasets and settings used. We use the same super network for training and evaluation in each task.

In practice, different hyperparameters are often tuned with a validation set for different tasks according to different datasets and losses. For example, HRNet~\cite{wang2020deep} is trained using two different settings for segmentation and keypoint estimation tasks.
In this work, we follow the common training settings~\cite{ding2021learning} of each task, \ie, the setting in HRNet~\cite{wang2020deep} for segmentation and keypoint estimation, AtomNAS~\cite{mei2020atomnas} for classification, and PointPillar~\cite{lang2019pointpillars} for 3D detection.

As for the choice of $\lambda$ for each task, we first empirically tuned it so that HR-NAS-A's FLOPs is comparable to the least FLOPs among the baseline models, then we relaxed the restriction so that HR-NAS-B reaches SOTA yet still costs less FLOPs than the best baseline models. Currently, the searched model size cannot be controlled precisely by $\lambda$. We will strengthen it by incorporating other techniques as our future work.
See below for details.

\noindent \textbf{ImageNet for Image Classification.}~~The ILSVRC 2012 classification
dataset~\cite{deng2009imagenet} consists of 1,000 classes, with a number of 1.2 million training images and 50,000 validation images.
Follow the common practice in ~\cite{tan2019mnasnet,you2020greedynas,mei2020atomnas,mingxing2019efficient,stamoulis2019single}, we adopt a RMSProp optimizer with momentum 0.9 and weight decay 1e-5; exponential moving average (EMA) with decay 0.9999; and exponential learning rate decay. 
The input size is $224 \times 224$.
The initial learning rate is set to 0.064 with batch size 1024 on 16 Tesla V100 GPUs for 350 epochs, and decays by 0.97 every 2.4 epochs.
By setting the coefficient of the L1 penalty term $\lambda$ to 1.8e-4 and 1.2e-4, we obtain our HR-NAS-A and HR-NAS-B.
Unless specified, we adopt the ReLU activation and the basic data augmentation scheme, i.e., random resizing and cropping, and random horizontal flipping, and use single-crop for evaluation.
For experiments of HR-NAS$\dag \ddag$, we also adopt the SE module~\cite{hu2018squeeze}, Swish activation~\cite{ramachandran2017searching}, and RandAugment~\cite{cubuk2020randaugment} for better performance.
We report the top-1 Accuracy as the evaluation metric.

\noindent \textbf{Cityscapes for Semantic Segmentation.}~~The Cityscapes dataset~\cite{cordts2016cityscapes} contains high-quality pixel-level annotations of 5000 images with size 1024x2048 (2975, 500, and 1525 for the training, validation, and test sets respectively) and about 20000 coarsely annotated training images. Following works~\cite{ding2020every,cordts2016cityscapes}, 19 semantic labels are used for evaluation without considering the void label.
In this work, the input size is set to $512 \times 1024$.
We use an AdamW optimizer with momentum 0.9 and weight decay 1e-5; exponential moving average (EMA) with decay 0.9999.
The initial learning rate is set to 0.04 with batch size 32 on 8 Tesla V100 GPUs for 430 epochs.
The learning rate and momentum follow the onecycle scheduler with a minimum learning rate of 0.0016.
By setting the coefficient of the L1 penalty term $\lambda$ to 1.6e-4 and 6.0e-5, we obtain our HR-NAS-A and HR-NAS-B.
We use a basic data augmentation, \ie, random resizing and cropping, random horizontal flipping, and photometric distortion for training and single-crop testing with a test size of $1024 \times 2048$.
We report the mean Intersection over Union (mIoU), mean (macro-averaged) Accuracy (mAcc), and overall (micro-averaged) Accuracy (aAcc) as the evaluation metrics.

\noindent \textbf{ADE20K for Semantic Segmentation.}~~The ADE20K dataset~\cite{zhou2017scene} contains 150 classes and diverse scenes with 1,038 image-level labels. The dataset is divided into 20K/2K/3K images for training, validation, and testing respectively.
In this work, the input size and testing size is set to $512 \times 512$ and $512 \times 2048$, respectively.
The model is trained with a batch size of 64 on 8 Tesla V100 GPUs for 200 epochs.
We use the same optimizer, learning rate scheduler, data augmentation, and penalty weight $\lambda$ as in the Cityscapes dataset.
We report the mean Intersection over Union (mIoU) as the evaluation metric.

\begin{figure*}[!htb]
    \centering
    \includegraphics[width=0.98\linewidth]{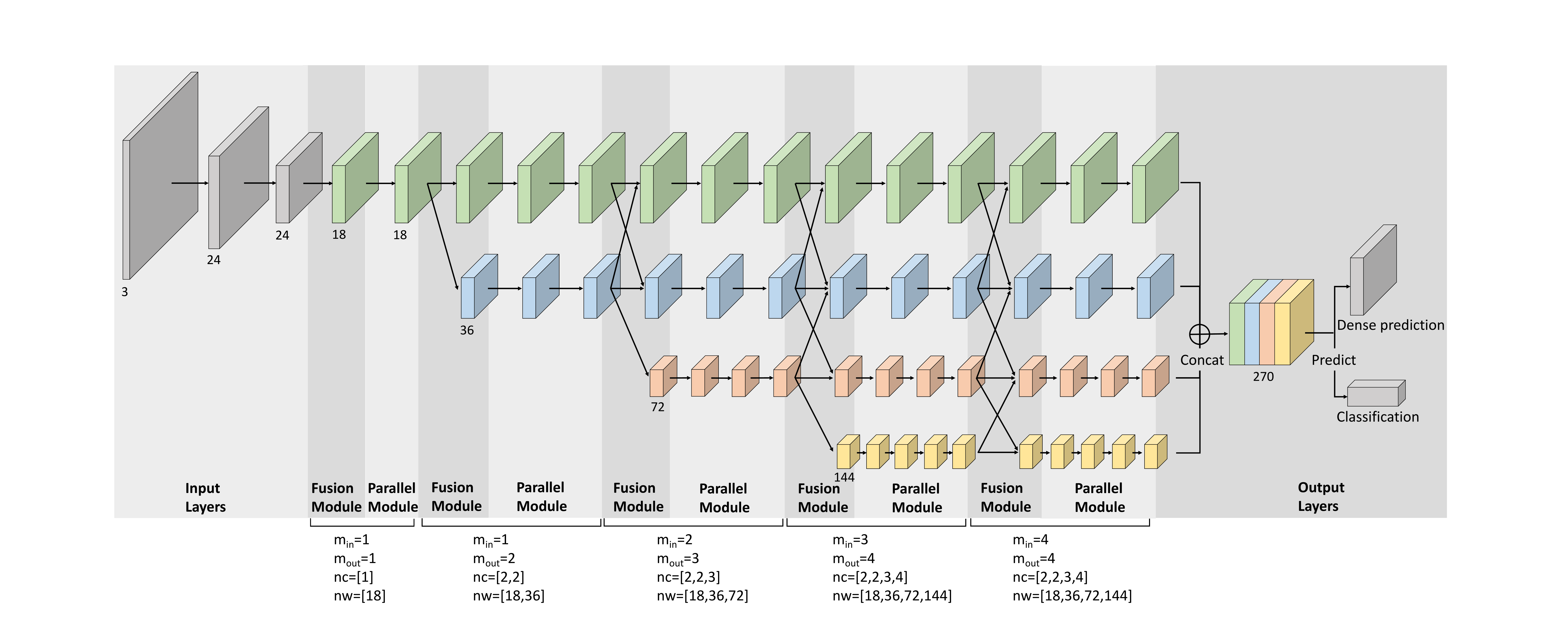}
    \vspace{-2pt}
    \caption{Visualization of our super network architecture. $m_{in}$ and $m_{out}$ denote the input and output numbers of branches in the fusion module. $nc$ and $nw$ denote the number of searching blocks and the number of channels in the parallel module, respectively.
    The arrows represent the \cells~and the cubes represent the feature maps.
    % The number on the arrow represents the index of the searching blocks in a branch ($nc_m, m \in {1,2,3,4}$) of parallel module. 
    The number under the cube represents the number of channels.
    }
    \label{fig:whole_hrnet}
    \vspace{-4pt}
\end{figure*}

\noindent \textbf{COCO Keypoint for Human Pose Estimation.}~~The COCO dataset~\cite{lin2014microsoft} contains over 
$200,000$ images and $250,000$ person instances labeled with $17$ keypoints. 
We train our model on the COCO \texttt{train2017} set, including $57K$ images and $150K$ person instances. We evaluate our approach on the \texttt{val2017}, containing $5000$ images.
In this work, we train the model using input sizes of $256 \times 192$ and $384 \times 288$ with batch size 384 and 192 on 8 Tesla V100 GPUs for 210 epochs, respectively.
Following HRNet~\cite{wang2020deep}, the initial learning rate is set to 1e-3 with a multistep scheduler (decayed by a factor of 0.1 in 170 and 200 epochs). We use an Adam optimizer with momentum 0.9 and weight decay 1e-8; exponential moving average (EMA) with decay 0.9999.
By setting the coefficient of the L1 penalty term $\lambda$ to 1e-6 and 1e-8, we obtain our HR-NAS-A and HR-NAS-B.
We use random scaling and rotation as only data augmentation for training and single-crop testing.
We report average precision (AP), recall scores (AR), $\text{AP}^M$ for medium objects, and $\text{AP}^L$ for large objects as evaluation metrics.

\noindent \textbf{KITTI for 3D Object Detection.}~~The KITTI 3D object detection dataset~\cite{geiger2012we} is widely used for monocular and LiDAR-based 3D detection. It consists of 7,481 training images and 7,518 test images as well as the corresponding point clouds and the calibration parameters, comprising a total of 80,256 2D-3D labeled objects with three object classes: Car, Pedestrian, and Cyclist. Each 3D ground truth box is assigned to one out of three difficulty classes (easy, moderate, hard) according to the occlusion and truncation levels of objects.
In this work, we follow the train-val split \cite{chen20153d}, which contains 3,712 training and 3,769 validation images. The overall framework is based on Pointpillars~\cite{lang2019pointpillars}. The input point points are projected into bird's-eye view (BEV) feature maps by a voxel feature encoder (VFE). The projected BEV feature maps ($496 \times 432$) are then used as input of our 2D network for 3D/BEV detection.
Following~\cite{lang2019pointpillars,ding2020learning}, we set, pillar resolution: 0.16m, max number of pillars: 12000, and max number of points per pillar: 100.
We use the onecycle scheduler with an initial learning rate of 2e-3, a minimum learning rate of 2e-4, and batch size 16 on 8 Tesla V100 GPUs for 80 epochs.
We use an AdamW optimizer with momentum 0.9 and weight decay 1e-2.
We apply the same data augmentation, \ie, random mirroring and flipping, global rotation and scaling, and global translation for 3D point clouds as in Pointpillar~\cite{lang2019pointpillars}.
At inference time, we apply axis-aligned nonmaximum suppression (NMS) with an overlap threshold of 0.5 IoU.
We report standard average precision (AP) as the evaluation metric.

\section{Network Architecture}
As shown in Fig.~\ref{fig:whole_hrnet}, we visualize our entire super network used in all experiments. It begins with two $3\times 3$ convolutions with stride 2 and number of channels 24, which are followed by five parallel modules (respectively with 1, 2, 3, 4, 4 branches); a fusion module is inserted between every two adjacent parallel modules, to obtain multi-scale features. The numbers of channels for the four branches in parallel modules are 18, 36, 72, 144, respectively.

\section{Ablative Results for Transformer}
In this section, we conduct two ablative experiments to study the impact of the projection size $s$, the encoder-decoder structure, and the attention mechanism on the performance of our lightweight Transformer. For both experiments, we take the searched network on Multi-branch + MixConv space (without Transformer) in Tab.6 of the main paper as a strong baseline.

\begin{table}[t]
    \caption{Comparisons of different projection size $s$ of Transformer on the CityScapes validation set. The query number $n$ is set to 8.}
    \centering
    {\footnotesize
    \scalebox{1}{ 
    \tabcolsep=0.14cm
    \begin{tabular}{l|r|r|ccc}
        \toprule
        \multirow{1}{*}{Input size} & Params & FLOPs & \textbf{mIoU}(\%) & mACC(\%) & aACC(\%) \\
        \midrule
    Baseline & 1.120M & 1.863G & 71.99 & 80.33 & 95.40 \\
	$2\times2$ & 1.180M & 1.863G & 72.27 & 80.74 & 95.40 \\
	$4\times4$ & 1.246M & 1.864G & 73.32 & 81.76 & 95.45 \\
	$8\times8$ & 2.273M & 1.872G & \textbf{74.22} & \textbf{82.36} & \textbf{95.52} \\
	$16\times16$ & 18.543M & 1.969G & 74.18 & 82.07 & 95.50 \\
% 	32x32 & 278.233M & 3.455G &  \\
    \bottomrule
    \end{tabular}
    }}
    \label{tab:spatial_size}
    \vspace{-6pt}
\end{table}

\noindent\textbf{Projection Sizes.}~~We evaluate our Transformers with different projected spatial sizes $s$. From Tab.~\ref{tab:spatial_size} we can see that when $s$ goes from 0 to 8, the mIoU keeps increasing at the expense of small extra cost (\ie, FLOPs). Further increasing $s$ brings no gain in performance but drastically increasing FLOPs. We therefore choose $s=8$ throughout the experiments.

\begin{table}[t]
    \caption{Ablation study of our lightweight Transformer with $n=8$ and $s=8$ on the CityScapes validation set. Notations:
    `Enc' -- only the encoder of Transformer is used, `Enc + Dec' -- both the encoder and decoder are used in Transformer, `channel' -- use each channel as a token, `spatial' -- use each spatial position as a token.
    }
    \centering
    {\footnotesize
    \scalebox{0.93}{ 
    \tabcolsep=0.1cm
    \begin{tabular}{l|r|r|ccc}
        \toprule
        \multirow{1}{*}{Input size} & Params & FLOPs & \textbf{mIoU}(\%) & mACC(\%) & aACC(\%) \\
        \midrule
	Baseline & 1.120M & 1.863G & 71.99 & 80.33 & 95.40 \\
	SE~\cite{hu2018squeeze} & 2.101M & 1.864G & 72.81 & 81.33 & 95.35 \\
	Non-local~\cite{wang2018non} & 1.317M & 2.951G & 72.50 & 81.32 & 95.34 \\
	Enc (spatial) & 1.184M & 1.866G & 72.61 & 80.97 & 95.26 \\
	Enc (channel) & 1.723M & 1.869G & 73.66 & 82.10 & 95.50 \\
	Enc + Dec (spatial) & 1.204M & 1.867G & 73.54 & 81.87 & 95.44 \\
	Enc + Dec (channel) & 2.273M & 1.872G & \textbf{74.22} & \textbf{82.36} & \textbf{95.52} \\
    \bottomrule
    \end{tabular}
    }}
    \label{tab:transformer}
  \vspace{-9pt}
\end{table}

\noindent\textbf{Attention Structures and Mechanisms.}~~We also conduct ablative experiments to validate the effectiveness of our Transformer. We discuss (1) encoder-decoder structures and (2) two kinds of attention mechanisms by transposing the feature, \ie, `channel' – use each channel of the flattened feature map as a token, `spatial' – use each spatial position as a token.
As shown in Tab.~\ref{tab:transformer}, our Transformer obtains the best performance when both encoder and decoder are used on channel-wise tokens.
Our Transformer also significantly outperforms its counterparts such as SE~\cite{hu2018squeeze} and Non-local~\cite{wang2018non} on dense prediction tasks. Since the channel-wise lightweight transformer shows better performance, we set it as the default in this work.

\section{Visualization of Visual Recognition Results}

\begin{figure}[!htb]
  \centering
    \begin{minipage}[b]{0.49\linewidth}
      \centering
      \includegraphics[width=\linewidth]{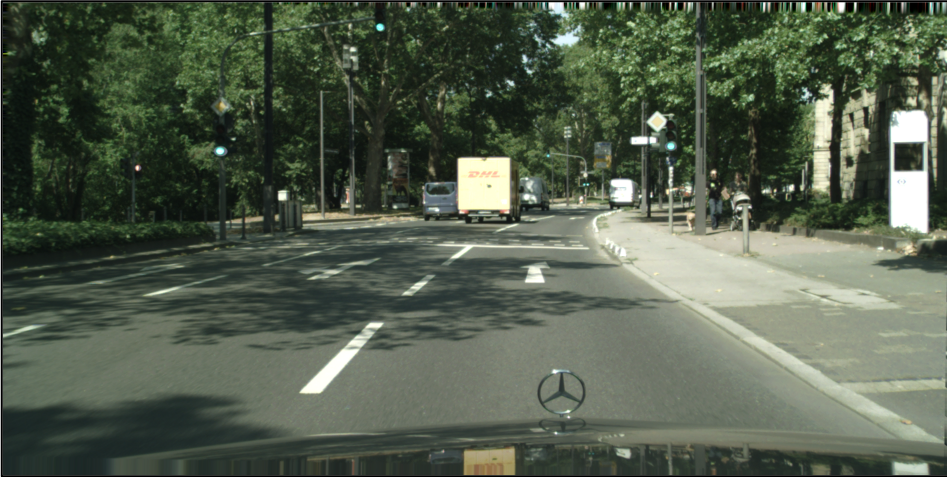}\vspace{2pt}
      \includegraphics[width=\linewidth]{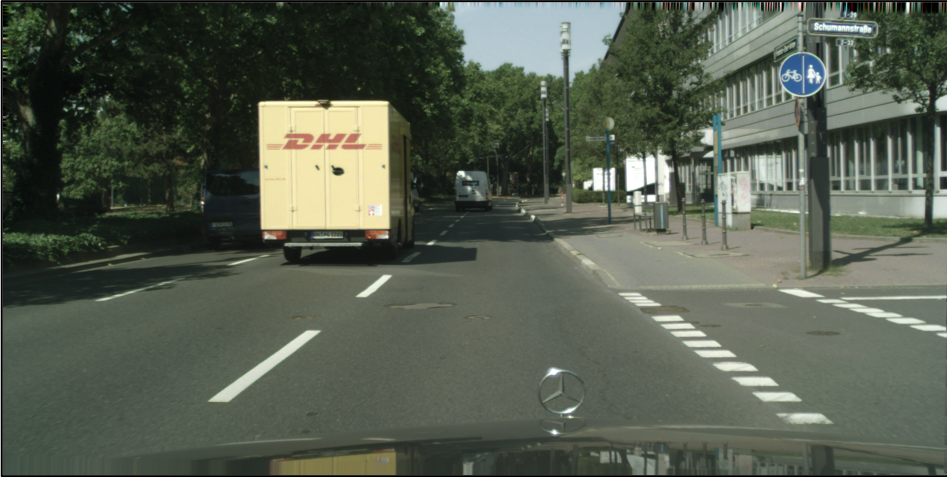}\vspace{2pt}
      \includegraphics[width=\linewidth]{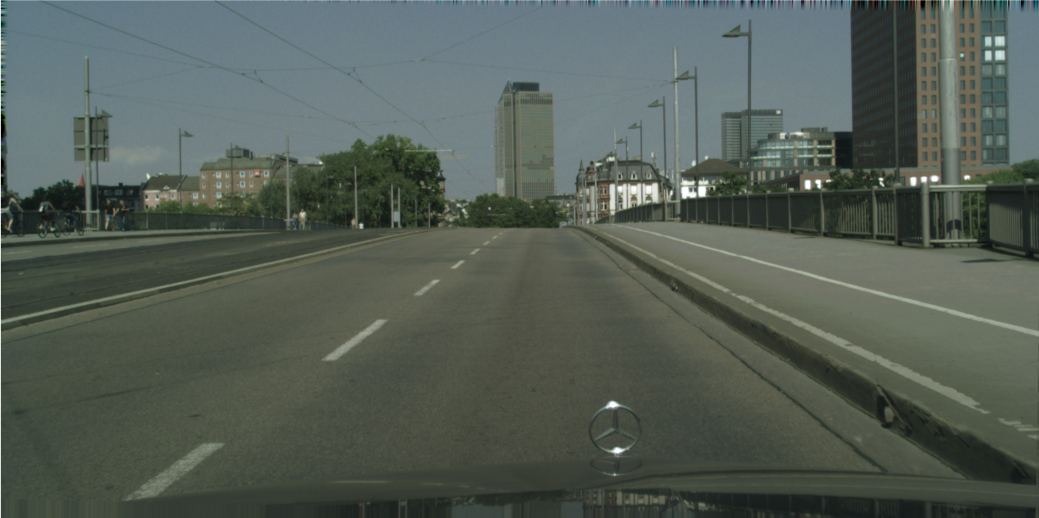}\vspace{2pt}
      \includegraphics[width=\linewidth]{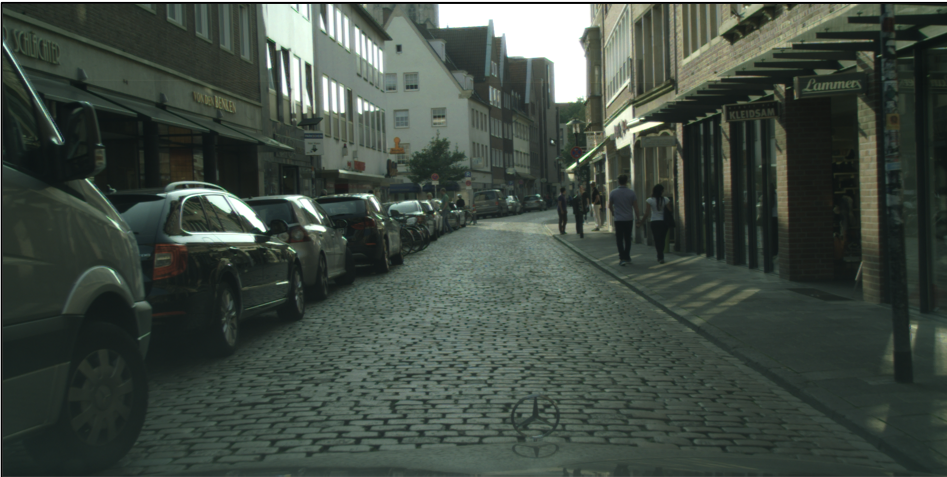}
    \end{minipage}
  \hfill
    \begin{minipage}[b]{0.49\linewidth}
      \centering
      \includegraphics[width=\linewidth]{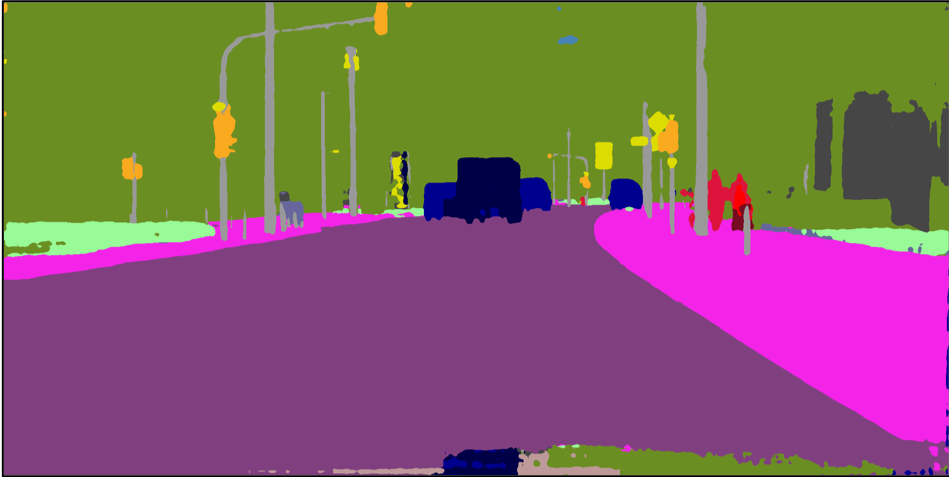}\vspace{2pt}
      \includegraphics[width=\linewidth]{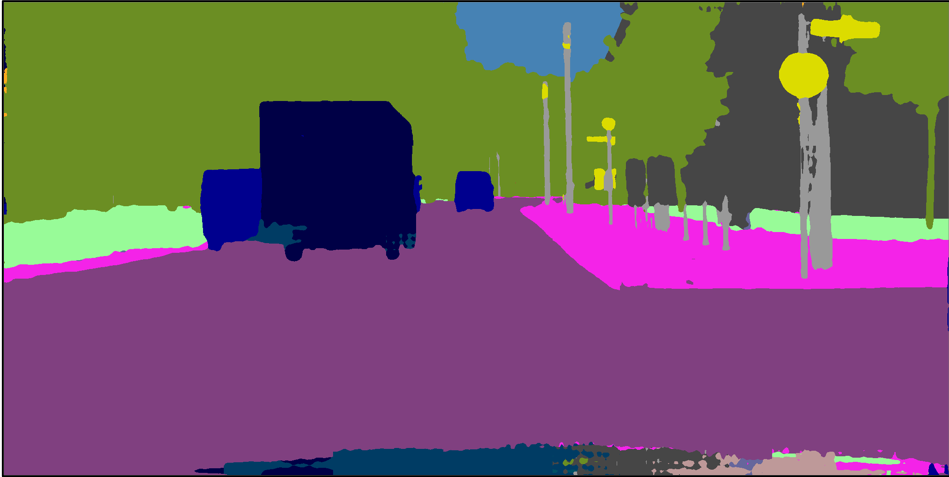}\vspace{2pt}
      \includegraphics[width=\linewidth]{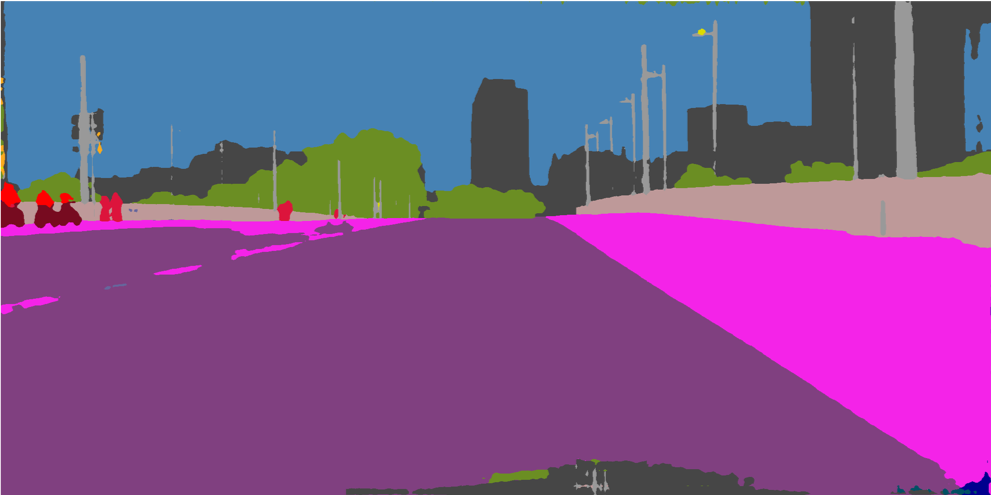}\vspace{2pt}
      \includegraphics[width=\linewidth]{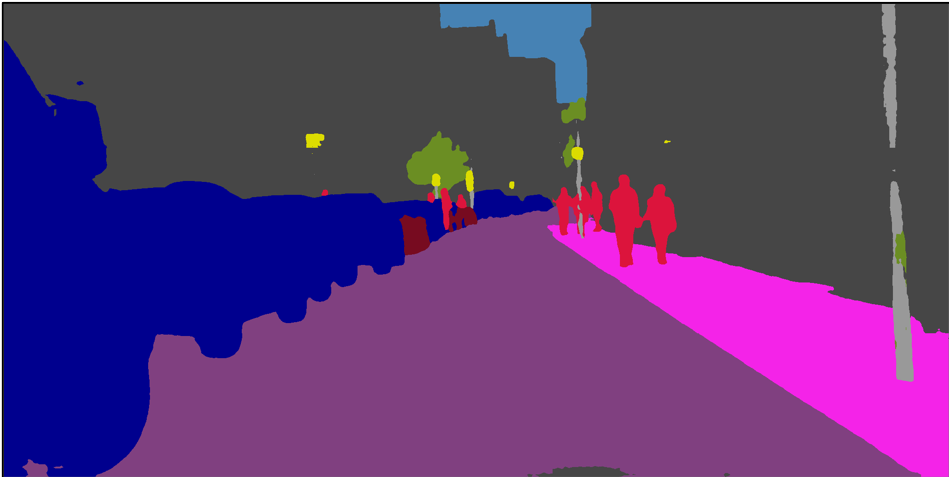}
    \end{minipage}
  \caption{Visualization of semantic segmentation results (left: original images; right: segmentation maps) on Cityscapes.}
  \label{fig:vis_seg}
\end{figure}

\begin{figure}[!htb]
  \centering
    \begin{minipage}[b]{0.45\linewidth}
      \centering
      \includegraphics[width=\linewidth]{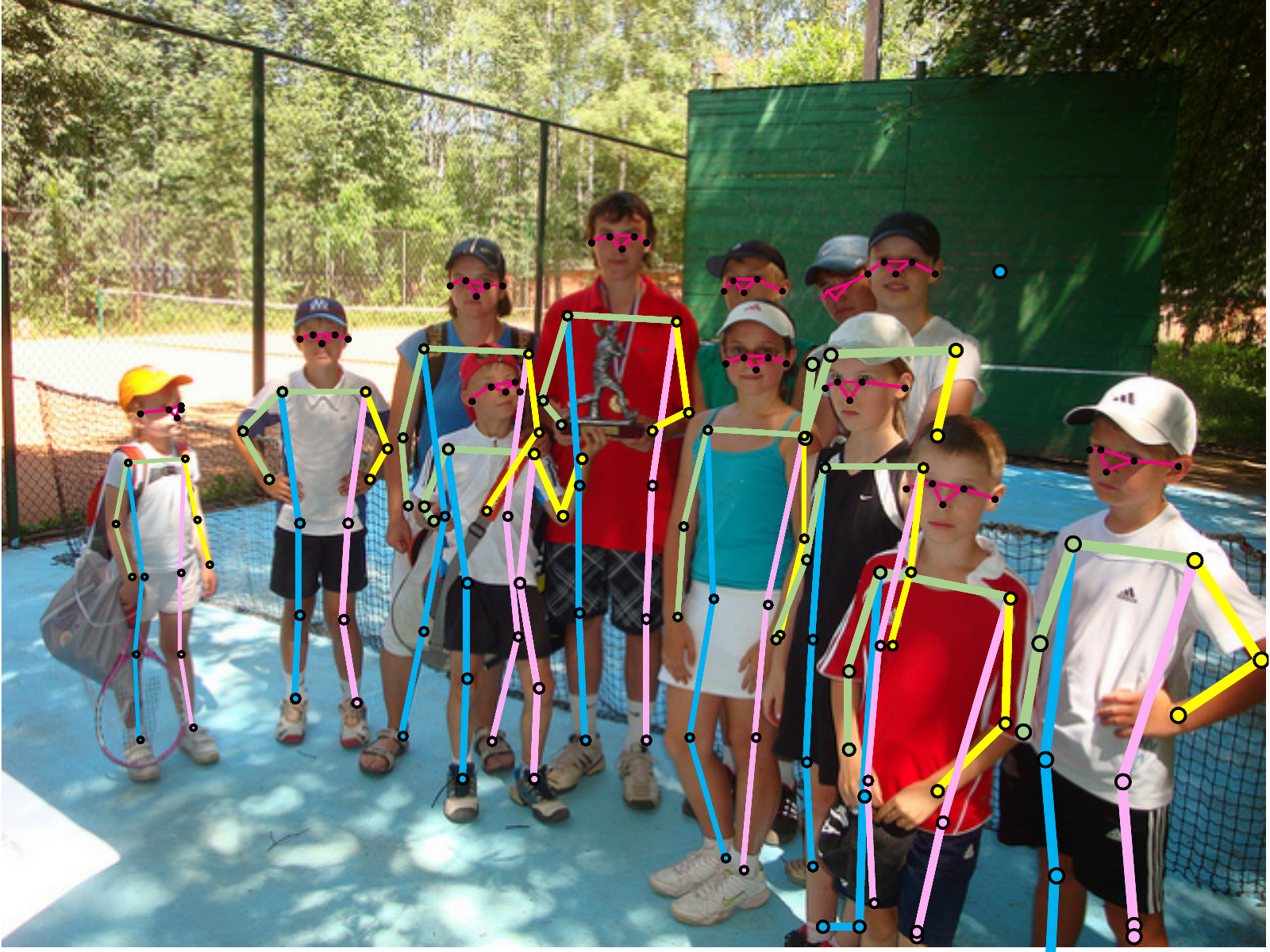}\vspace{2pt}
      \includegraphics[width=\linewidth]{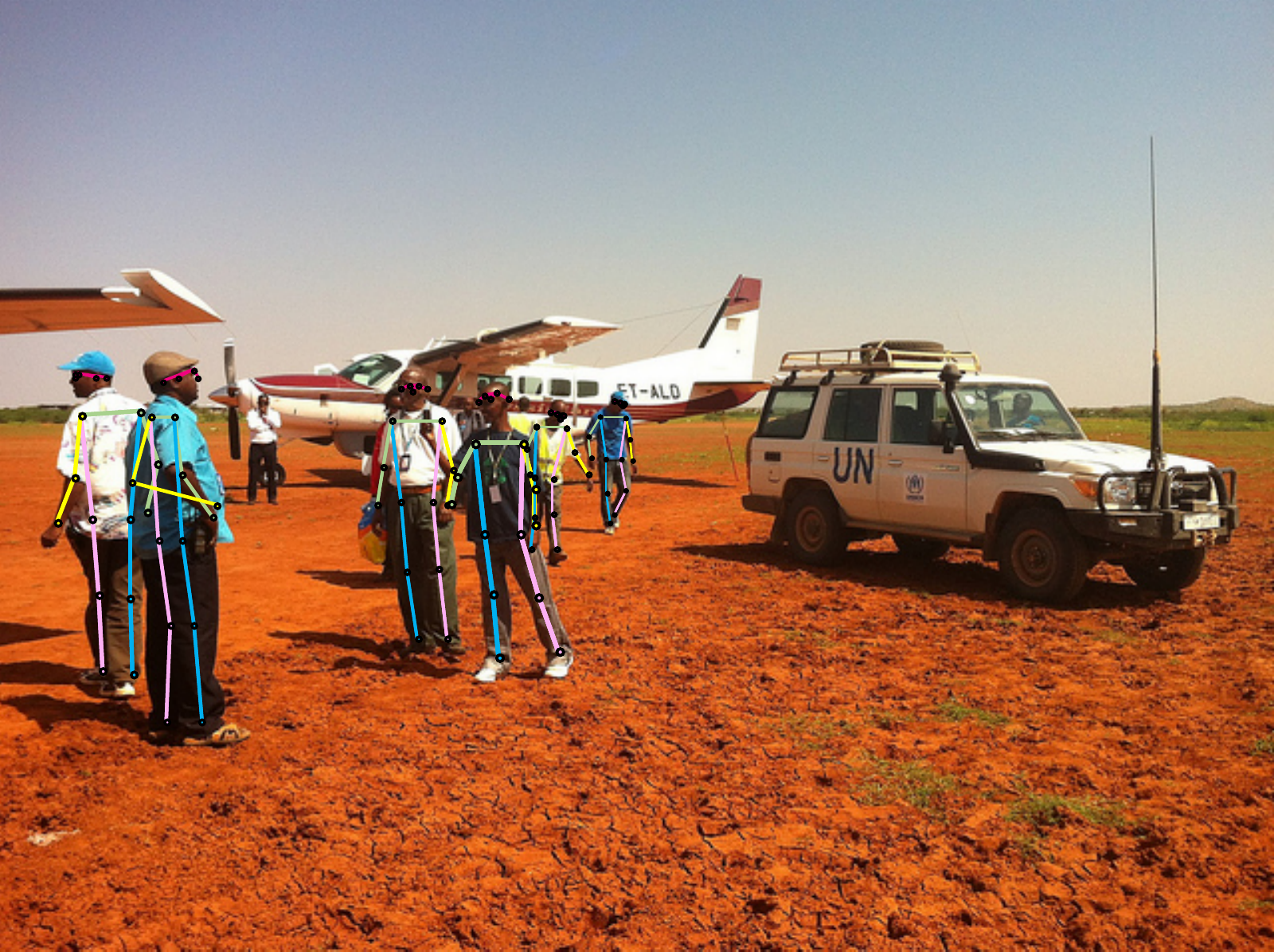}\vspace{2pt}
      \includegraphics[width=\linewidth]{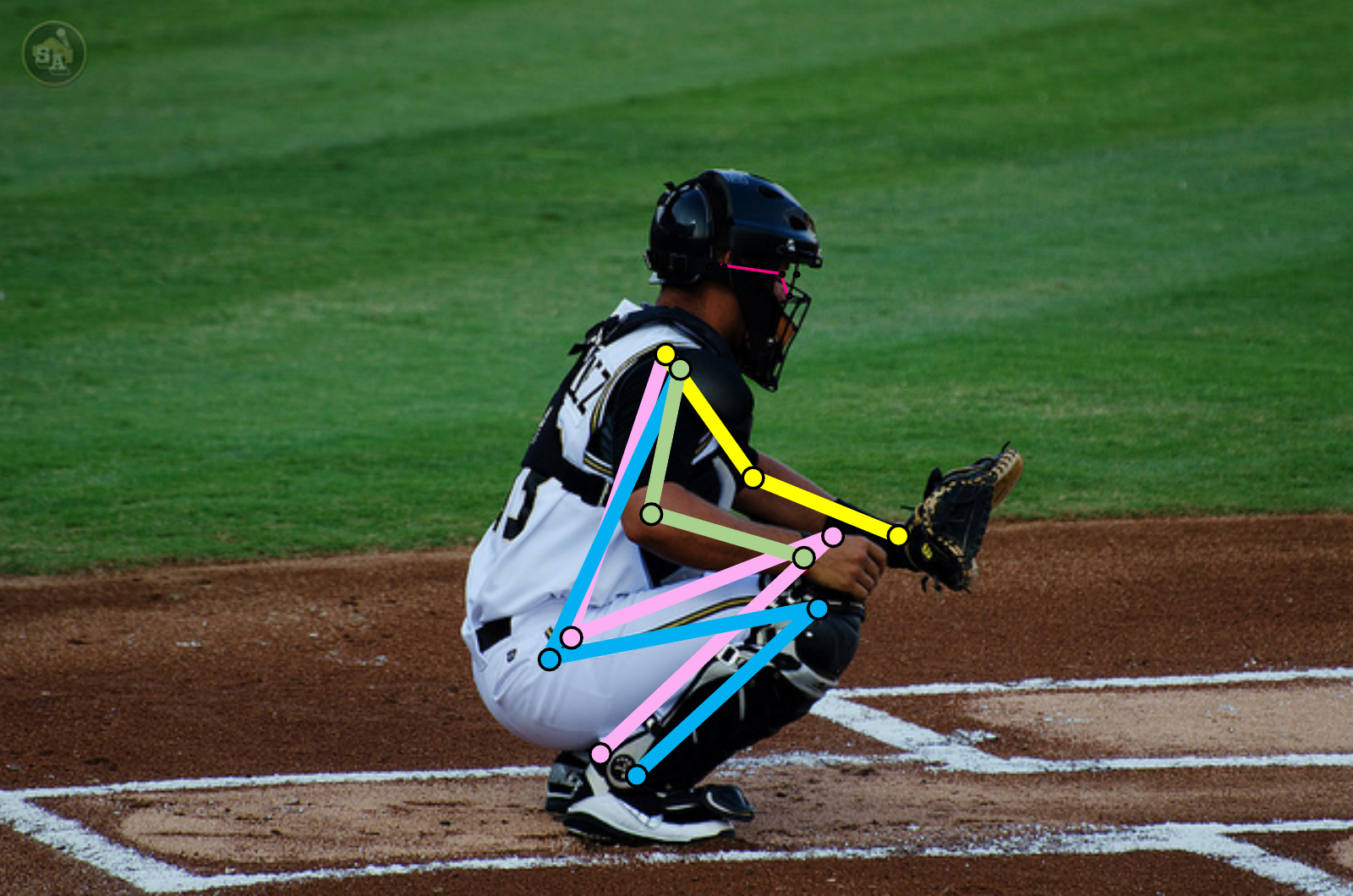}\vspace{2pt}
    \end{minipage}
%   \hfill
    \begin{minipage}[b]{0.45\linewidth}
      \centering
      \includegraphics[width=\linewidth]{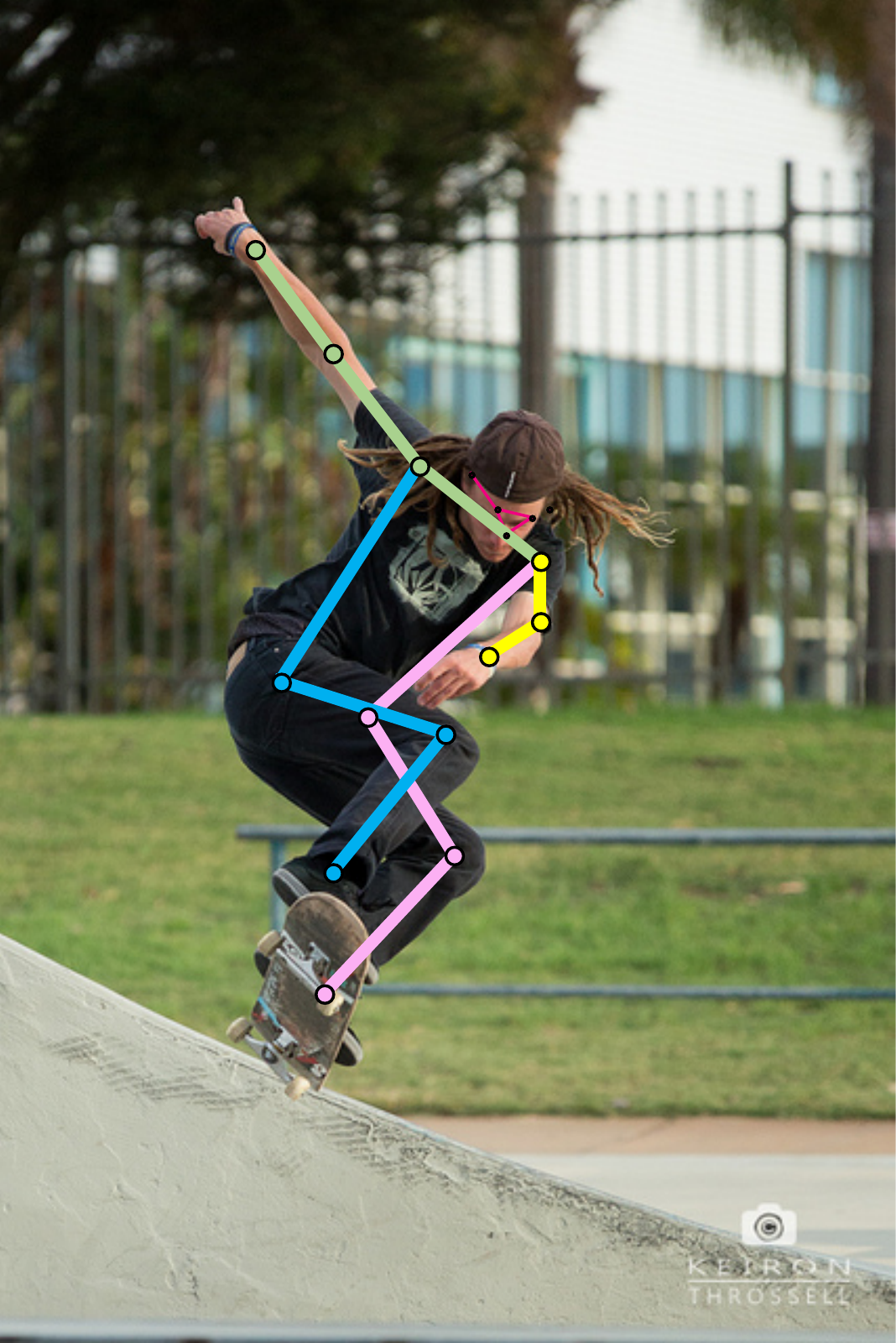}\vspace{4pt}
      \includegraphics[width=\linewidth]{figures/supp/13291.pdf}\vspace{2pt}
    \end{minipage}
  \caption{Visualization of human pose estimation on COCO.}
  \label{fig:vis_coco}
  \vspace{-6pt}
\end{figure}

\begin{figure}[!htb]
  \centering
    \begin{minipage}[b]{0.45\linewidth}
      \centering
      \includegraphics[width=\linewidth]{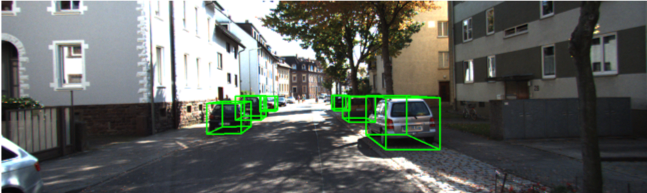}\vspace{2pt}
      \includegraphics[width=\linewidth]{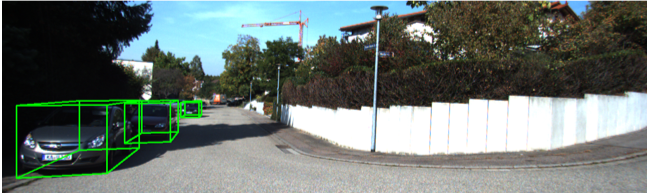}\vspace{2pt}
      \includegraphics[width=\linewidth]{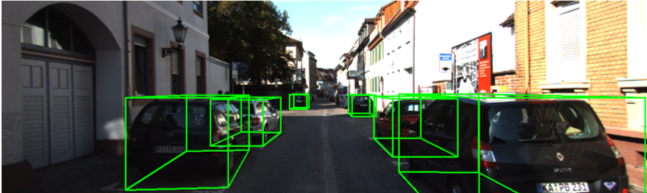}
    \end{minipage}
%   \hfill
    \begin{minipage}[b]{0.45\linewidth}
      \centering
      \includegraphics[width=\linewidth]{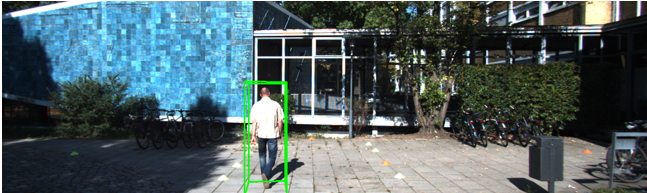}\vspace{2pt}
      \includegraphics[width=\linewidth]{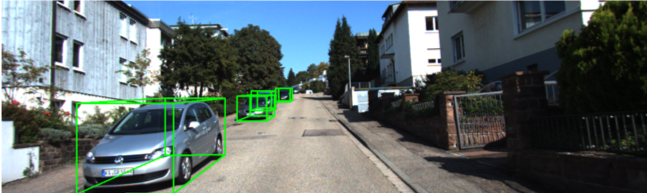}\vspace{2pt}
      \includegraphics[width=\linewidth]{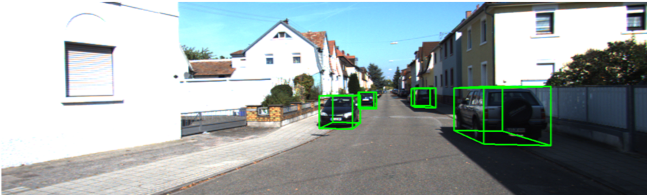}
    \end{minipage}
  \caption{Visualization of 3D object detection results on KITTI.}
  \label{fig:vis_3d}
\end{figure}

We visualize the results of HR-NAS-A on segmentation, human pose estimation, and 3D detection (Fig.~\ref{fig:vis_seg},~\ref{fig:vis_coco},~\ref{fig:vis_3d}).

\end{appendices}

{\small
\bibliographystyle{ieee_fullname}
\bibliography{egbib}
}

\end{document}